\documentclass[5p,twocolumn,times]{elsarticle}
\usepackage{lineno,hyperref}
\usepackage{amssymb}
\usepackage{bm}
\usepackage{amsmath}
\usepackage{subfigure}
\usepackage{algorithm, algorithmic}
\usepackage{multirow}
\usepackage{epstopdf}
\usepackage{booktabs}
\usepackage {setspace}

\makeatletter
\renewcommand{\@thesubfigure}{\hskip\subfiglabelskip}
\makeatother

\begin{document}
\begin{frontmatter}
\title{Towards the Desirable Decision Boundary by Moderate-Margin Adversarial Training}
\author[address1]{Xiaoyu Liang}
\ead{212009701006@zust.edu.cn}
\author[address1]{Yaguan Qian\corref{mycorrespondingauthor}}
\cortext[mycorrespondingauthor]{Corresponding author}
\ead{qianyaguan@zust.edu.cn}
\author[address1]{Jianchang Huang}
\ead{212109701007@zust.edu.cn}
\author[address2]{Xiang Ling}
\ead{lingxiang@iscas.ac.cn}
\author[address3]{Bin Wang}
\ead{wangbin2@hikvision.com}
\author[address4]{Chunming Wu}
\ead{wuchunming@zju.edu.cn}
\author[address5]{Wassim Swaileh}
\ead{wassim.swaileh@cyu.fr}

\address[address1]{School of Science, Zhejiang University of Science and Technology, China}
\address[address2]{Institute of Software, Chinese Academy of Sciences, China}
\address[address3]{The Zhejiang Key Laboratory of Multidimensional Perception Technology, Application and Cybersecurity, China}
\address[address4]{The College of Computer Science, Zhejiang University, China }
\address[address5]{IRISA Laboratory UMR 6074, INSA De Rennes, CNRS}

\begin{abstract}
Adversarial training, as one of the most effective defense methods against adversarial attacks, tends to learn an inclusive decision boundary to increase the robustness of deep learning models. However, due to the large and unnecessary increase in the margin along adversarial directions, adversarial training causes heavy cross-over between natural examples and adversarial examples, which is not conducive to balancing the trade-off between robustness and natural accuracy. In this paper, we propose a novel adversarial training scheme to achieve a better trade-off between robustness and natural accuracy. It aims to learn a moderate-inclusive decision boundary, which means that the margins of natural examples under the decision boundary are moderate. We call this scheme Moderate-Margin Adversarial Training (MMAT), which generates finer-grained adversarial examples to mitigate the cross-over problem. We also take advantage of logits from a teacher model that has been well-trained to guide the learning of our model. Finally, MMAT achieves high natural accuracy and robustness under both black-box and white-box attacks. On SVHN, for example, state-of-the-art robustness and natural accuracy are achieved.
\end{abstract}


\end{frontmatter}

\section{Introduction}
Deep neural networks (DNNs) have achieved great success in many tasks, including image classification; however, recent research indicates that DNNs are extremely vulnerable to adversarial attacks. The so-called adversarial attack is to mislead a classifier by adding some well-designed perturbations to a clean input. Although several countermeasures have been proposed \cite{a15,a16,a17}, adversarial training has confirmed to be one of the most effective methods. Nevertheless, the higher Robust Accuracy (RA) of adversarial training is often accompanied by more Natural Accuracy (NA) degradation, which prompts us to achieve a good trade-off between RA and NA.

\begin{figure}[t]
\setlength{\abovecaptionskip}{0.1cm} 
\setlength{\belowcaptionskip}{-0.4cm} 
\centering
\subfigure[(a)]{
\begin{minipage}[t]{0.27\linewidth}
\centering
\includegraphics[width=1in]{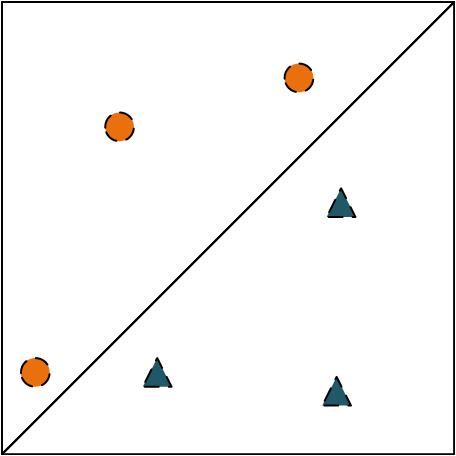}
\end{minipage}
}
\subfigure[(b)]{
\begin{minipage}[t]{0.27\linewidth}
\centering
\includegraphics[width=1in]{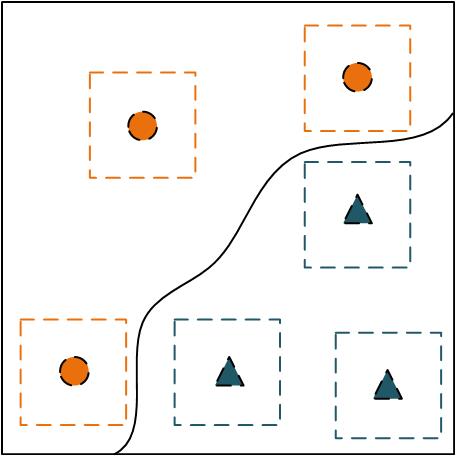}
\end{minipage}
}
\subfigure[(c)]{
\begin{minipage}[t]{0.26\linewidth}
\centering
\includegraphics[width=1in]{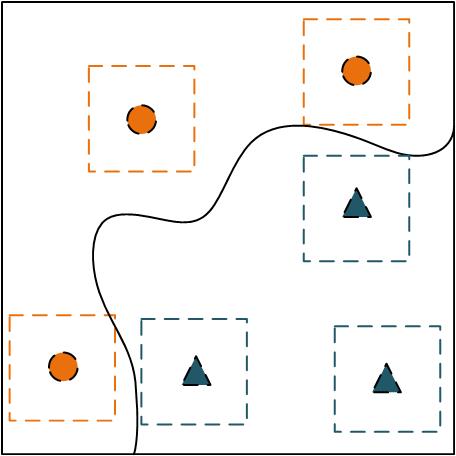}

\end{minipage}
}
\caption{A conceptual illustration of decision boundaries learned via (a) natural training, (b) standard adversarial training, and (c) moderate-margin adversarial training, respectively. The adversarially trained model needs a significantly more complicated decision boundary to memorize adversarial examples of training data.}
\label{fig1}
\end{figure}

In fact, although adversarially trained models have learned inclusive decision boundaries that keep adversarial counterparts within the original class, the unwarranted increase in the margin along certain adversarial directions is harmful, as shown in Fig. \ref{fig1}(c). The decision boundary's over-inclusiveness causes a heavy cross-over between natural and adversarial examples, which is not conducive to balancing the RA-NA trade-off. In this paper, we expect to alleviate the excessive adversarial directional margin. We begin our analysis by examining the decision boundaries of the DNNs obtained by different training methods. Next, we investigate the details of previous work \cite{a3,c11,c13} and discover that it mostly treats all examples equally and assigns uniform parameters to them. We then conduct experiments to confirm that this is one of the main causes of excessive directional margin. Thus, we propose a novel adversarial training scheme, namely Moderate-Margin Adversarial Training (MMAT).

Here we separate the training examples clearly, identifying both the critical difference between correctly classified and misclassified examples by a model as well as the characteristics of each correctly classified example. For misclassified examples, it does not make sense to study their robustness. Therefore, we consider a boosted loss function to better learn the examples themselves; for correctly classified examples, we design a new attack strategy based on the variability of their margins. It can significantly reduce the excessive adversarial directional margin for the examples near the decision boundary. In addition, we obtain a well-trained model for the same architecture and use it to guide the training process of our defense model, which can regulate the desirable course of the decision boundary and, to some extent, weaken the over-adjustment of the decision boundary along adversarial directions. Ultimately, we obtain a moderate-inclusive decision boundary shown in Fig. \ref{fig1}(b), which improves classification accuracy on natural examples without compromising robustness. To sum up, our main contributions are:
\begin{enumerate}[1)]
	\item We consider the RA-NA trade-off to be the issue of how broad the margins of natural examples should be;
    \item We propose a new adversarial defense method, MMAT, which generates finer-grained adversarial examples (FAEs) based on projected gradient descent (PGD), reducing the heavy cross-over between natural and adversarial examples;
    \item Extensive experiments on CIFAR-10 and SVHN show that MMAT has a higher NA while keeping a competitive RA for adversarial examples compared with other adversarial training methods.
\end{enumerate}
\vspace{-0.6cm}

\section{Related work}
\paragraph{\bf{Adversarial attacks}}Szegedy et al. \cite{a1} first observed that DNNs are highly vulnerable to adversarial examples. Goodfellow et al. \cite{a2} proposed Fast Gradient Sign Method (FGSM), which makes use of the gradient of the model to generate adversarial examples. Madry et al. \cite{a3} proposed Projected Gradient Descent (PGD) method as a universal "first-order adversary", i.e., the most active attack utilizing the local first-order information about the network. Moreover, boundary-based methods like DeepFool \cite{a4} and optimization-based methods like CW \cite{a5} were also developed, making adversarial defense more challenging. Recently, the ensemble of diverse attack methods - auto-attack \cite{a6} by Croce et al., consisting of APGD-CE \cite{a6}, APGD-DLR \cite{a6}, FAB \cite{a7}, and Square Attack \cite{a8}, became a popular benchmark for testing the robustness of models. These are white-box attacks since the adversary has full knowledge of DNNs. Numerous works have also been published that investigate the transferability of adversarial examples for black-box attacks. Liu et al. \cite{a9} were the first to study the transferability of targeted adversarial examples. Dong et al. \cite{a10} showed that iterative attack methods incorporating the momentum term achieved better transferability. Xie et al. \cite{a11} boosted the transferability of adversarial examples by creating diverse input patterns with random resizing and random padding.

\paragraph{\bf{Adversarial training}}Many of the early adversarial defenses \cite{a12,a13,a14} are broken by stronger attacks. Among various defense strategies, adversarial training is currently the best defense method against adversarial attacks. The initial idea of adversarial training is first brought to light by \cite{a1}, where neural networks are trained on a mixture of adversarial examples and clean examples. Goodfellow et al. \cite{a2} went further and proposed FGSM to produce adversarial examples during training. Mathematically, adversarial training is formulated as a minimax optimization problem. Madry et al. \cite{a3} employed a multi-step gradient-based attack known as PGD attack for solving the inner problem. Wang et al. \cite{c1} investigated the influence of misclassified examples and proposed MART to further improve robustness. However, it is often observed that these methods compromise the accuracy of unperturbed examples. Such observations had led prior work to speculate on a trade-off between the two fundamental notions of natural accuracy and robustness. Zhang et al. \cite{c11} regularized the output from natural images and adversarial examples with the KL divergence function and proposed TRADES, which can trade natural accuracy off against adversarial robustness. Zhang et al. \cite{c18} proposed searching for the least adversarial data for AT, which could be called FAT. Cui et al. \cite{c14} proposed LBGAT to adapt the logits of one model trained on clean data to guide adversarial training. Rade et al. \cite{c13} incorporated additional wrongly labelled examples during training and presented HAT to provide a notable improvement in accuracy without compromising robustness.
\section{Problem statement}
\subsection{Preliminaries}
Suppose $\mathbb{D}=\{(\bm x_i,y_i)\}_{i=1}^N$ is a dataset where $x_i\in \mathbb{R}^d$ is a natural example, and $y_i\in {1,2,\cdots,K}$ is its corresponding label. A DNN $f_{\bm \theta}$ is a function with parameter $\bm \theta$ to predict an input example $\bm x_i$: $f_{\bm \theta}(\bm x_i)={\operatorname {arg\,max}}_{k=1,\cdots,K} \bm z_k(\bm x_i)$, $\bm p_k(\bm x_i)={e^{\bm z_k(\bm x_i)}}/{\sum_{k^{'}=1}^{K } e^{\bm z_k(\bm x_i)}}$, where $\bm z_k(\bm x_i)$ is the logits output of the network with respect to class $k$, and $\bm p_k(\bm x_i)$ is the probability (softmax on logits) belonging to class $k$. An adversarial example $ {\bm x_i}'$ is crafted by adding a small perturbation $\bm \delta_i$ to $\bm x_i$, which misleads DNNs as $f_{\bm \theta}( {\bm x_i}')\neq y_i$, while $f_{\bm \theta}( {\bm x_i}')=y_i$. Small perturbation means that $\bm x_i$ satisfy $\{ {\bm x_i}'\in \mathbb{B}(\bm x_i,\epsilon):\| {\bm x_i}'-\bm x_i\leq\epsilon\| \}$, where $\|\cdot\|$ is a generic distance norm that can be $\|\cdot\|_\infty$ and $\|\cdot\|_2$.
\paragraph{\bf{Decision boundary}}Given a classifier $f_{\bm \theta}$, a dataset $\mathbb{D}={(\bm x_i,y_i)}_{i=1}^N$, we can define the decision boundary $\mathbb D'\subseteq \mathbb R^d$ of $f_{\bm \theta}$ as the set:

\hspace{-0.4mm}
\begin{equation}\label{eq1}
 \mathbb D':=\{\bm x_i \in \mathbb R^d|\exists l\neq j, \bm p_l(\bm x_i)=\bm p_j(\bm x_i)=\max_k\bm p_k(\bm x_i)\}
\end{equation}

\paragraph{\bf{Margin}}Given a unit vector $\bm v\in \mathbb S^{d-1}$, the margin $m$ at $\bm x_i$ along the direction $\bm v$ is given by:

\begin{equation}\label{eq2}
m(\bm x_i, \bm v) = \underset {a} {\operatorname {arg\,min}} \mid a\mid \quad s.t. \bm x_i+a\bm v \in \mathbb D'
\end{equation}
\paragraph{\bf{Adversarial directions}}Given a classifier $f_{\bm \theta}$, a dataset $\mathbb{D}={(\bm x_i,y_i)}_{i=1}^N$, we define the set of adversarial directions as $V= \{{\bm v_i}/{\|\bm v_i\|_2}\}_{i=1}^{N}$ where $\bm v_i$ is abtained by solving:

\begin{align}\label{eq3}
\bm v_i = \max_{\bm \delta:\|\bm \delta\|_{p\leq\epsilon}} L(f_{\bm \theta}(\bm x_i+\bm \delta),y_i)
\end{align}where $L(\cdot)$ is an arbitrary loss function. This optimization problem is usually solved via PGD.

\begin{figure*}[t]
\setlength{\abovecaptionskip}{-0.3cm}
\setlength{\belowcaptionskip}{-0.4cm}
\centering
\subfigure[]{
\begin{minipage}[t]{0.22\linewidth}
\centering
\includegraphics[width=1.5in]{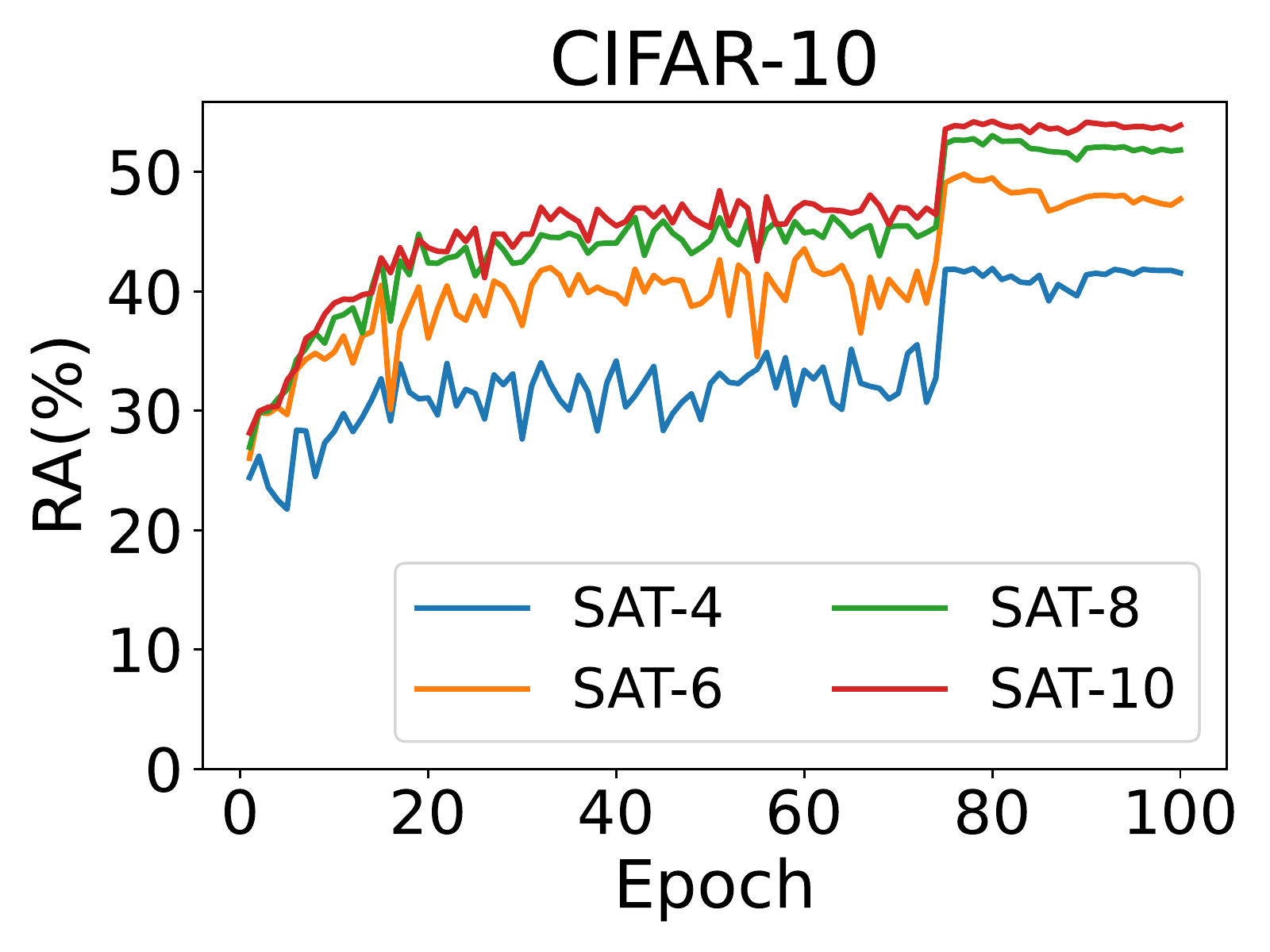}

\end{minipage}
}
\subfigure[]{
\begin{minipage}[t]{0.22\linewidth}
\centering
\includegraphics[width=1.5in]{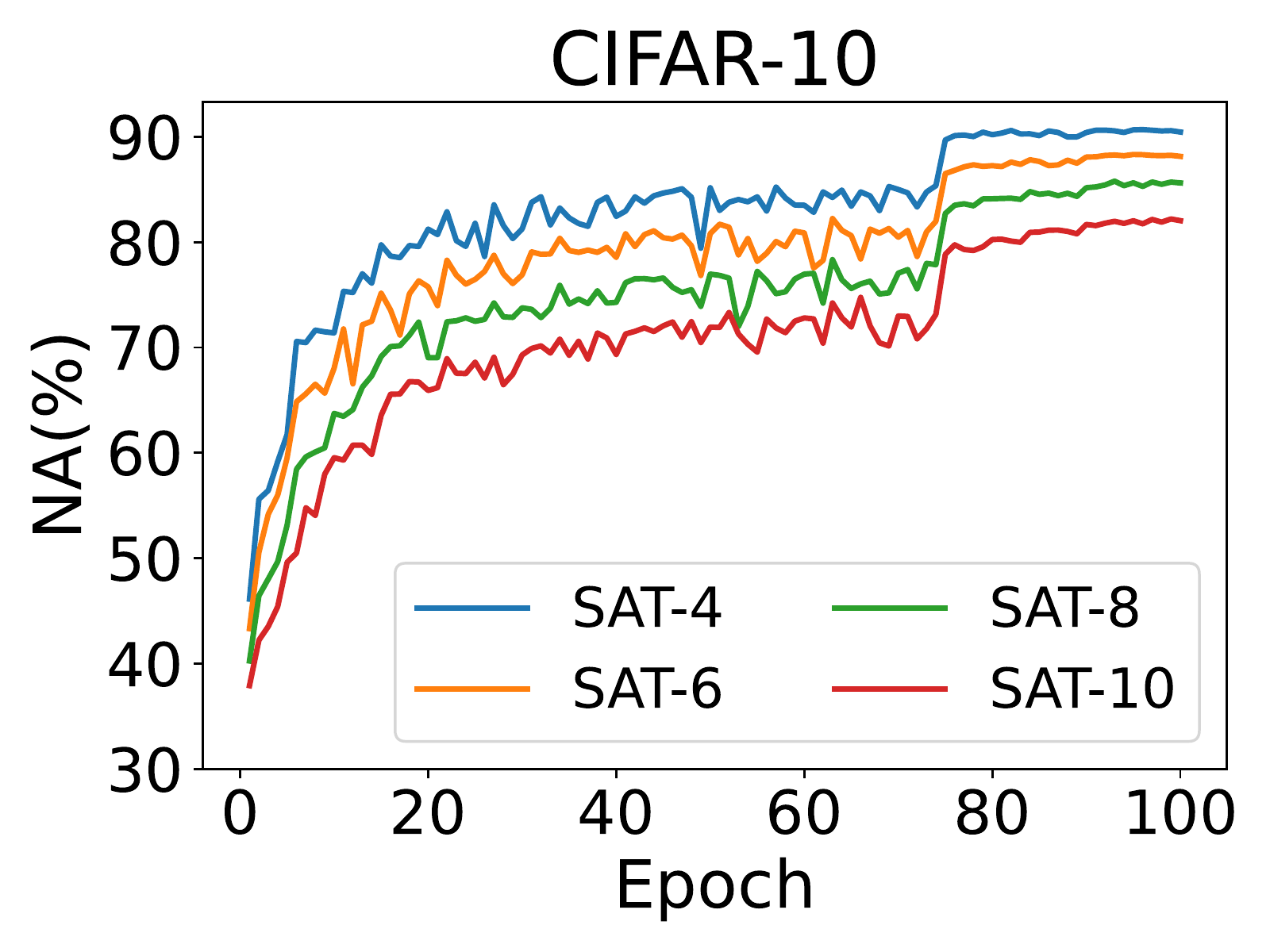}
\end{minipage}
}
\subfigure[]{
\begin{minipage}[t]{0.22\linewidth}
\centering
\includegraphics[width=1.5in]{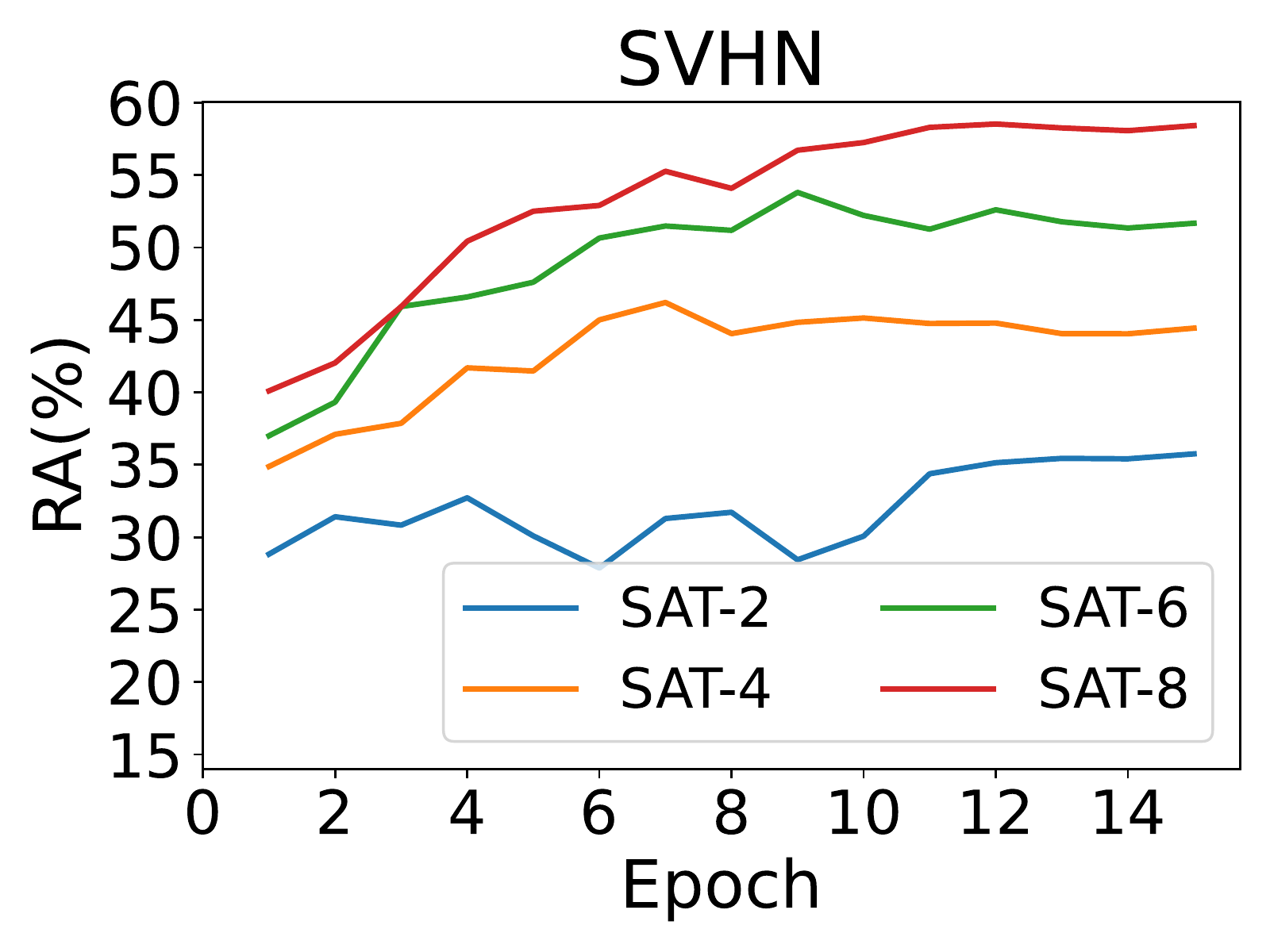}
\end{minipage}
}
\subfigure[]{
\begin{minipage}[t]{0.22\linewidth}
\centering
\includegraphics[width=1.5in]{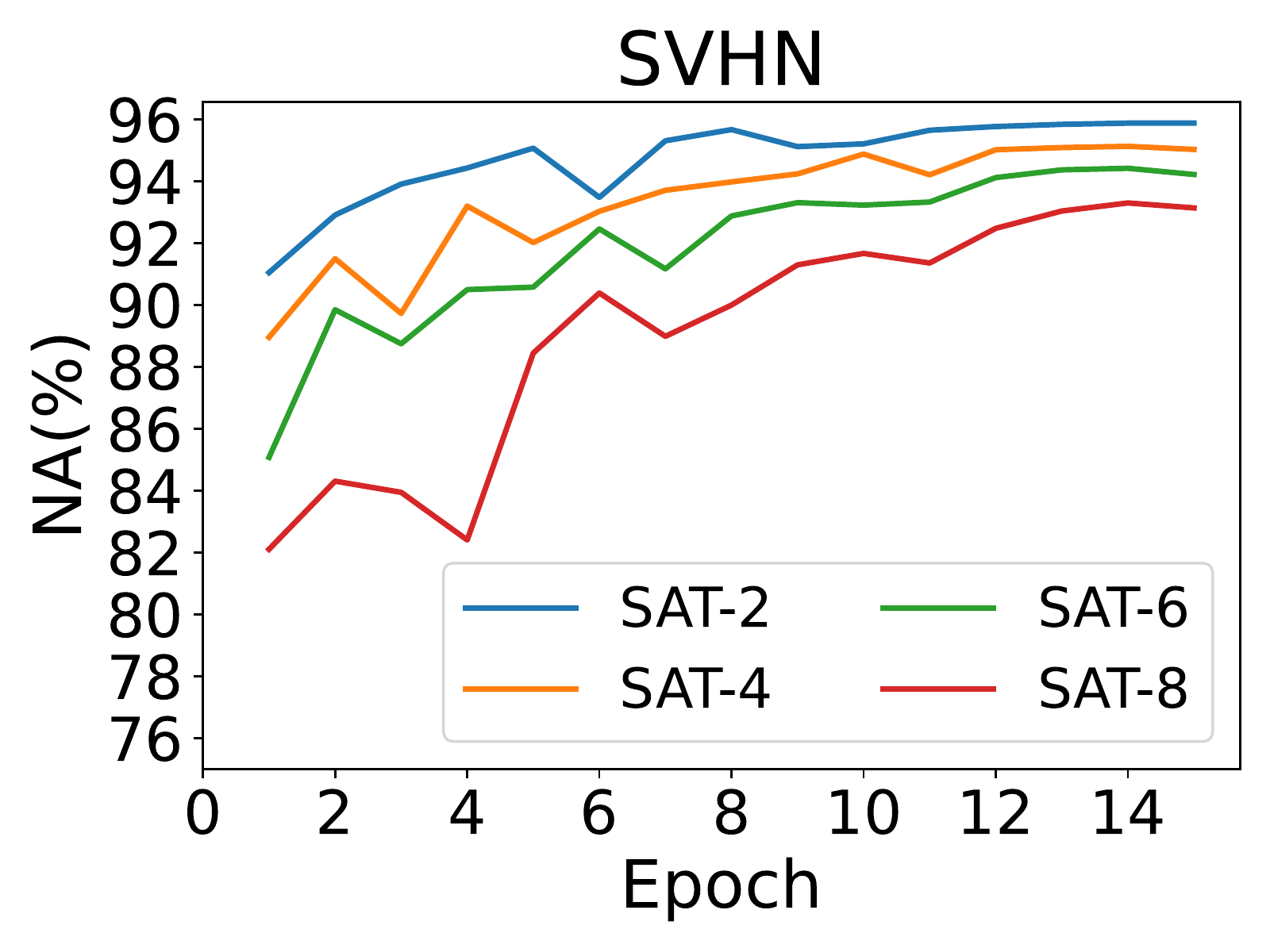}
\end{minipage}
}
\caption{The RA and NA of standard adversarially trained models. A trained model with $\epsilon=k/255$ is denoted by SAT-$k$. Here we train ResNet-18 on CIFAR-10 and PreActResNet-18 on SVHN. During training, we generate adversarial examples by using 10 steps of size $\epsilon/4$, where $\epsilon$ is the perturbation budget, and different $\epsilon$ represents different attack strengths.}
\label{fig2}
\end{figure*}
\subsection{Motivation}\label{set3.2}
A naturally trained model can attain perfect NA, but its average margin has a clear downward trend over the training process. Eventually, a large portion of the training examples as well as test examples lie close to the decision boundary after training\cite{c26}. The relatively low average margins lead to the vulnerability of DNNs to adversarial examples. In the adversarial setting, even though a slight downward trend is inevitable, it is not as severe as that in the natural setting, and this downward trend starts in later epochs compared to natural training. The adversarially trained models with larger average margins are robust while triggering a superfluous increase in the margin along adversarial directions as compared to the nominal increase required to attain robustness \cite{c13}. The excessive directional margin might be the reason for a reduction in natural accuracy. This argument is in line with the previous works by \cite{c31}, \cite{c32}.

We then obtain several robust models using the standard adversarial training (SAT) method \cite{a3}, and further investigate the differences in RA and NA among them. As shown in Fig. \ref{fig2}, the RA grows as $\epsilon$ increases, but the gain of this growth becomes gradually smaller. Meanwhile, the process is accompanied by a continued drop in NA. We realize that one reason for this phenomenon is that all examples share a uniform $\epsilon$. With the increase of $\epsilon$, more examples have excessive directional margin, especially those near the decision boundary, which significantly cross over the decision boundary and are located in the area of natural examples, causing a serious cross-over mixture problem \cite{c18}. Therefore, we prefer designing finer-grained parameters $\epsilon_i$ for different examples to alleviate the problem and exploit the potential of larger perturbation budgets.

Moreover, we can stimulate a slight push to the decision boundary during adversarial training to reduce the potential excess directional margin. This can be achieved by fine-tuning the robust model using the decision boundary learned by a natural model. To implement the idea, we consider a $weak$ robust model as our teacher model, such as SAT-6 in Fig. \ref{fig2}, which inherits the decision boundary of the natural model better in the process of improving robustness.
\begin{algorithm}[t]
\renewcommand{\algorithmicrequire}{\textbf{Input:}}
\renewcommand{\algorithmicensure}{\textbf{Output:}}
	\caption{The configuration of $\epsilon_i$}
	\label{alg1}
	\begin{algorithmic}[1]
		\REQUIRE A dataset $\mathbb D$, a strategy network $f_s$
		\ENSURE $\epsilon_i$
		\STATE Train a strategy network $f_s$ on $\mathbb{D}={(\bm x_i,y_i)}_{i=1}^N$
		\STATE $\mathbb D_A \leftarrow \varnothing$, $\mathbb D_B \leftarrow \varnothing$, $\mathbb D_C \leftarrow \varnothing$,
		\WHILE {$f_s(\bm x_i) \neq y_i$}
		\STATE $\epsilon_i \leftarrow 0$
        \ENDWHILE
		\WHILE {$f_s(\bm x_i) = y_i$}
        \STATE calculating the margin of all examples:\\ $M=\{m(\bm x_i)\}_{i=1}^{N}$
        \IF {$m(\bm x_i)\leq M_{P_{40}}$}
        \STATE $\mathbb D_A \leftarrow \mathbb D_A \cup \{\bm x_i\}$
        \STATE $\epsilon_i \leftarrow \epsilon_A = \max\limits_{m} m(\bm x_i),\bm x_i \in \mathbb D_A $
        \ELSIF {$m(\bm x_i)\leq M_{P_{70}}$}
        \STATE $\mathbb D_B \leftarrow \mathbb D_B \cup \{\bm x_i\}$
        \STATE $\epsilon_i \leftarrow \epsilon_B = \frac{1}{|\mathbb D_B|} \sum_{\bm x_i \in \mathbb D_B}m(\bm x_i)$
        \ELSIF {$m(\bm x_i)\leq M_{P_{100}}$}
        \STATE $\mathbb D_C \leftarrow \mathbb D_C \cup \{\bm x_i\}$
        \STATE $\epsilon_i \leftarrow \epsilon_C = \min\limits_{m} m(\bm x_i),\bm x_i \in \mathbb D_C $
        \ENDIF
        \ENDWHILE
	\end{algorithmic}
\end{algorithm}

\vspace{-2mm}
\section{Moderate-Margin Adversarial Training}
\vspace{-1mm}
To learn a moderate-inclusive decision boundary, we propose a novel adversarial training method named MMAT. This method achieves a good trade-off by designing two loss terms $L_1$ and $L_2$ respectively for adversarial and natural examples, and design a new attack strategy to generate our finer-grained adversarial examples.

\begin{algorithm}
	\renewcommand{\algorithmicrequire}{\textbf{Input:}}
	\renewcommand{\algorithmicensure}{\textbf{Parameter:}}
	\caption{Moderate-Margin Adversarial Training}
	\label{alg2}
	\begin{algorithmic}[1]
		\REQUIRE Training dataset $\mathbb{D}=\{(\bm x_i,y_i)\}_{i=1}^N$
		\ENSURE Batch-size $m$, learning rate $\eta$, scaling parameter $\lambda$, attack step size $\alpha$ and number of attack iterations $T$
		\STATE Train a teacher network $f_t$ on $\mathbb D$
		\STATE \textbf{repeat}
		 \STATE Read a mini-batch $\{(\bm x_{i_j},y_{i_j})\}_{j=1}^m$ from $\mathbb D$
		 \FOR {$j=1,2,\cdots,m$}
		  \STATE $ {\bm x_{i_j}}' \leftarrow \bm x_{i_j}+0.001 \times \mathcal N(\bm 0,\bm I) $, where $\mathcal N(\bm 0,\bm I)$ is the Gaussian distribution with zero mean and identity variance
		  \STATE Grade $\bm x_{i_j}$ based on its margin and set $\epsilon_{i_j}$ for it
		  \FOR {$t = 1,2,\cdots,T$}
		   \STATE $ {\bm x_{i_j}}' \leftarrow \prod_{\mathcal B(\bm x_{i_j},\epsilon_{i_j})}( {\bm x_{i_j}}' + \alpha \cdot sign\nabla_{ {\bm x_{i_j}}'}CE)$
          \ENDFOR
		\ENDFOR
        \STATE $\bm \theta \leftarrow \bm \theta - \frac{\eta}{m}\sum_{j=1}^{m}\nabla_{\bm \theta}(L_1+L_2/{\lambda})$
        \STATE \textbf{until} training completed
	\end{algorithmic}
\end{algorithm}
\subsection{Loss function}
The first term of loss function, $L_1$, is for the adversarial examples. Obviously, adversarial examples' classification is more arduous than natural examples', the decision boundary is no longer flat, and its local curvature profoundly affects the final performance of the defense model, as shown in Fig. \ref{fig1}, so the boosted cross entropy loss (BCE) is chosen here instead of the commonly used cross entropy loss. Thus $L_1$ is defined as follows:

\begin{align}\label{eq8}
\hspace{-3mm}
L_1= &-\log(\bm p_{y_i}(\bm {x_i}')) -\log(1-\max_{k\neq y_i}\bm p_k(\bm {x_i}'))
\end{align}
where $\bm p_k(\bm {x_i}')$ is the probability belonging to class $k$, the first term $-\log(\bm p_{y_i}(\bm {x_i}'))$ is the commonly used CE loss, denoted CE $(\bm p(\bm {x_i}'),y_i)$, and the second term $-\log(1-\max_{k\neq y_i}\bm p_k(\bm {x_i}'))$ is a margin term used to improve the classifier's decision margin, allowing it to be a robust classifier with high confidence.

The second term, $
L_{2}$, is for natural examples. The teacher model $f_t$ is a well-trained $weak$ robust model (like SAT with $\epsilon={6}/{255}$). The decision boundary of $f_{\bm \theta}$ will be fine-tuned in real time during training, and the fine-tuning will prevent the undesirable excessive rise in the margin to some extent, thus making it possible to achieve a better trade-off. We make use of logits from the teacher model to help reshape the decision boundary of our robust model, which reduces damage to the NA. Here we use MSE loss as $L_2$:

\begin{align}\label{eq9}
\hspace{-5mm}
L_2= \parallel \bm z^t(\bm x_i)-\bm z^{\bm \theta}(\bm x_i) \parallel_2^2
\end{align}
where $\bm z^{\bm \theta}(\bm x_i)$, $\bm z^t(\bm x_i)$ is logit output of input $\bm x_i$ of $f_{\bm \theta}$ and $f_t$ respectively. When the student $f_{\bm \theta}$ is trained with MSE loss, its representations attempt to follow the shape of the teacher’s representations. Besides, MSE is efficient in transferring the teacher's information to a student \cite{c29}.

\begin{figure*}[t]
\setlength{\abovecaptionskip}{0.2cm}
\setlength{\belowcaptionskip}{-0.4cm}
\centering
\subfigure[(a)]{
\begin{minipage}[t]{0.22\linewidth}
\centering
\includegraphics[width=1.5in,trim=0 0 0 0]{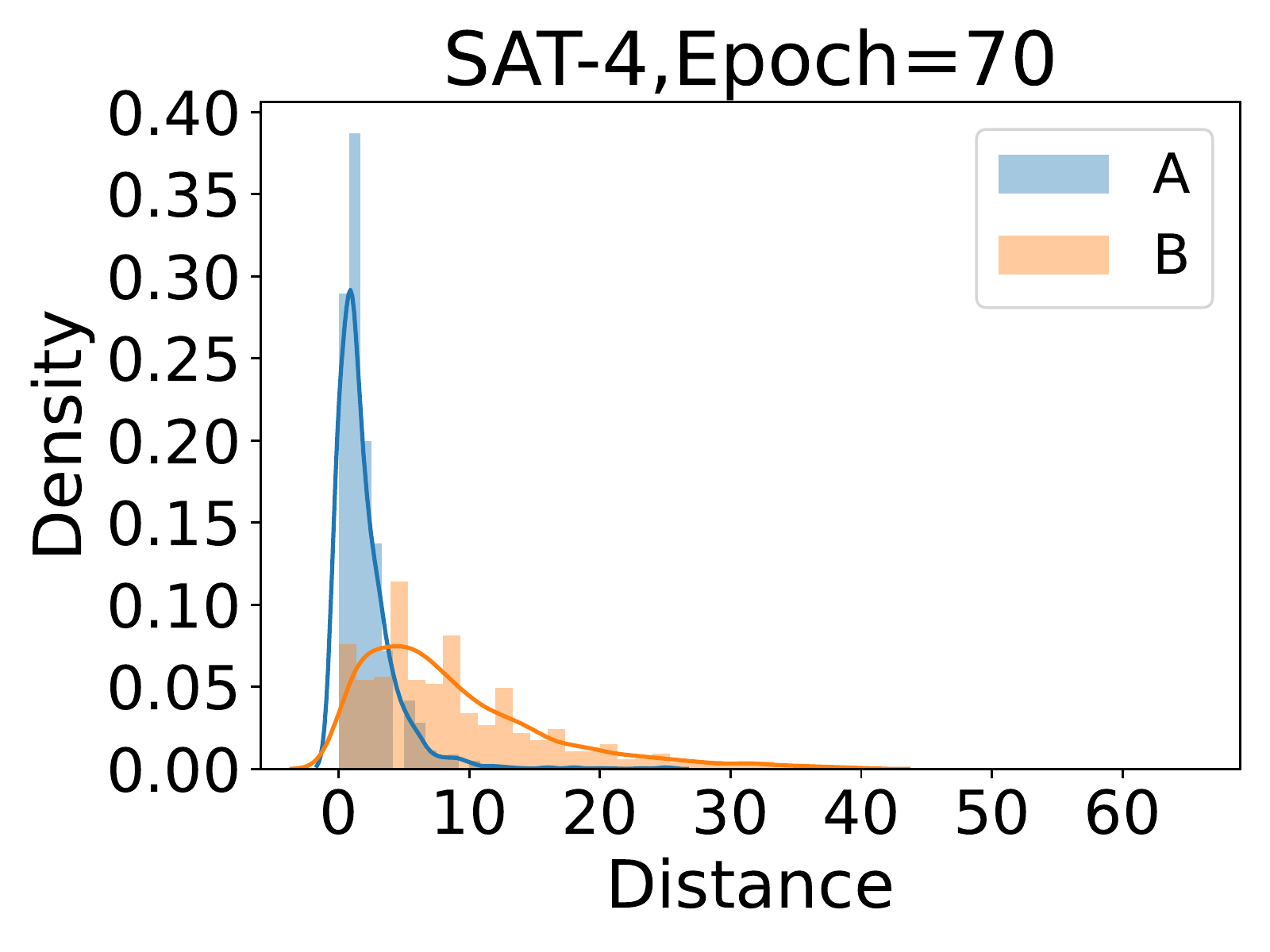}
\end{minipage}
}
\subfigure[(b)]{
\begin{minipage}[t]{0.22\linewidth}
\centering
\includegraphics[width=1.5in,trim=0 0 0 0]{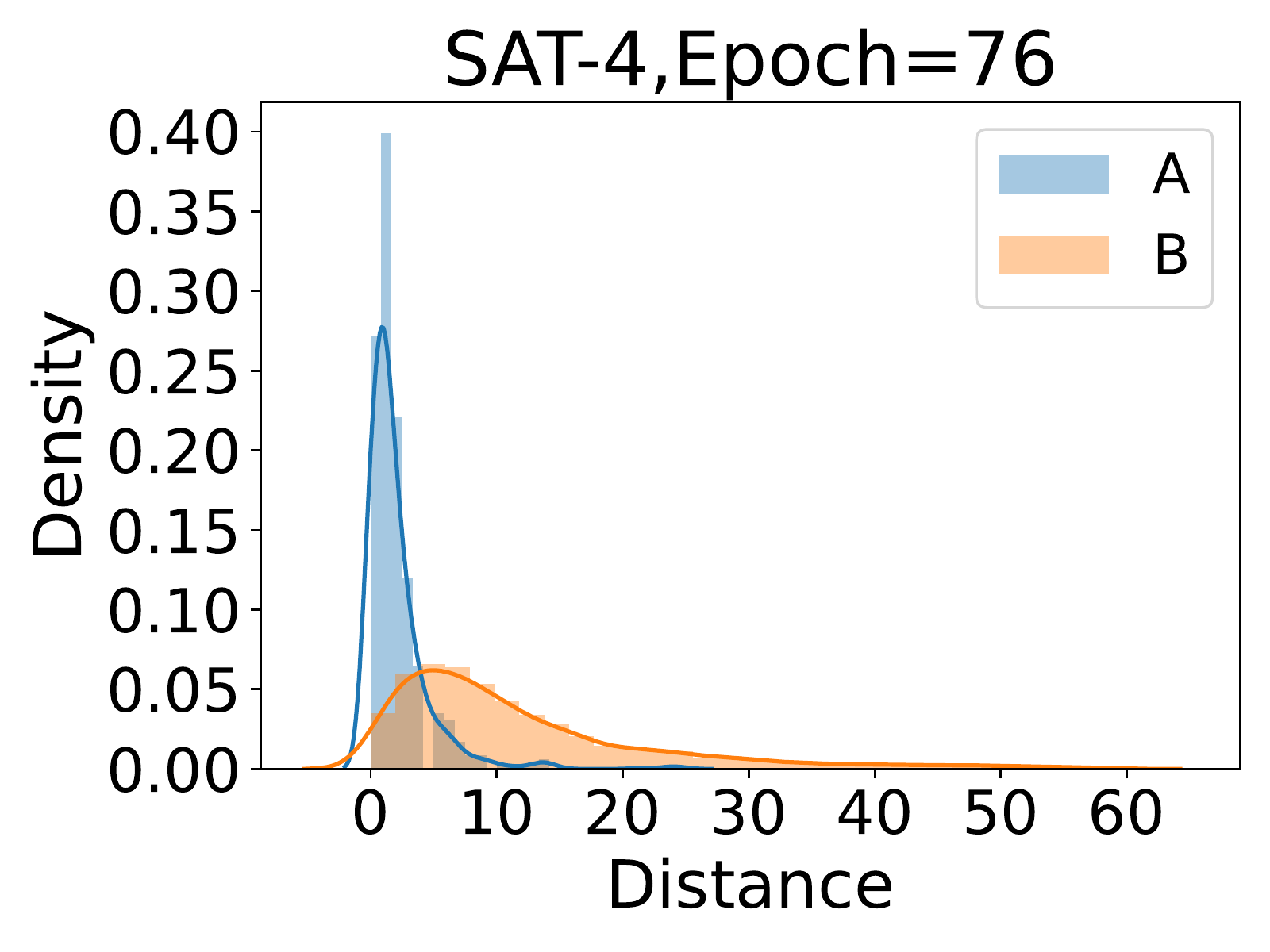}
\end{minipage}
}
\subfigure[(c)]{
\begin{minipage}[t]{0.22\linewidth}
\centering
\includegraphics[width=1.5in,trim=0 0 0 0]{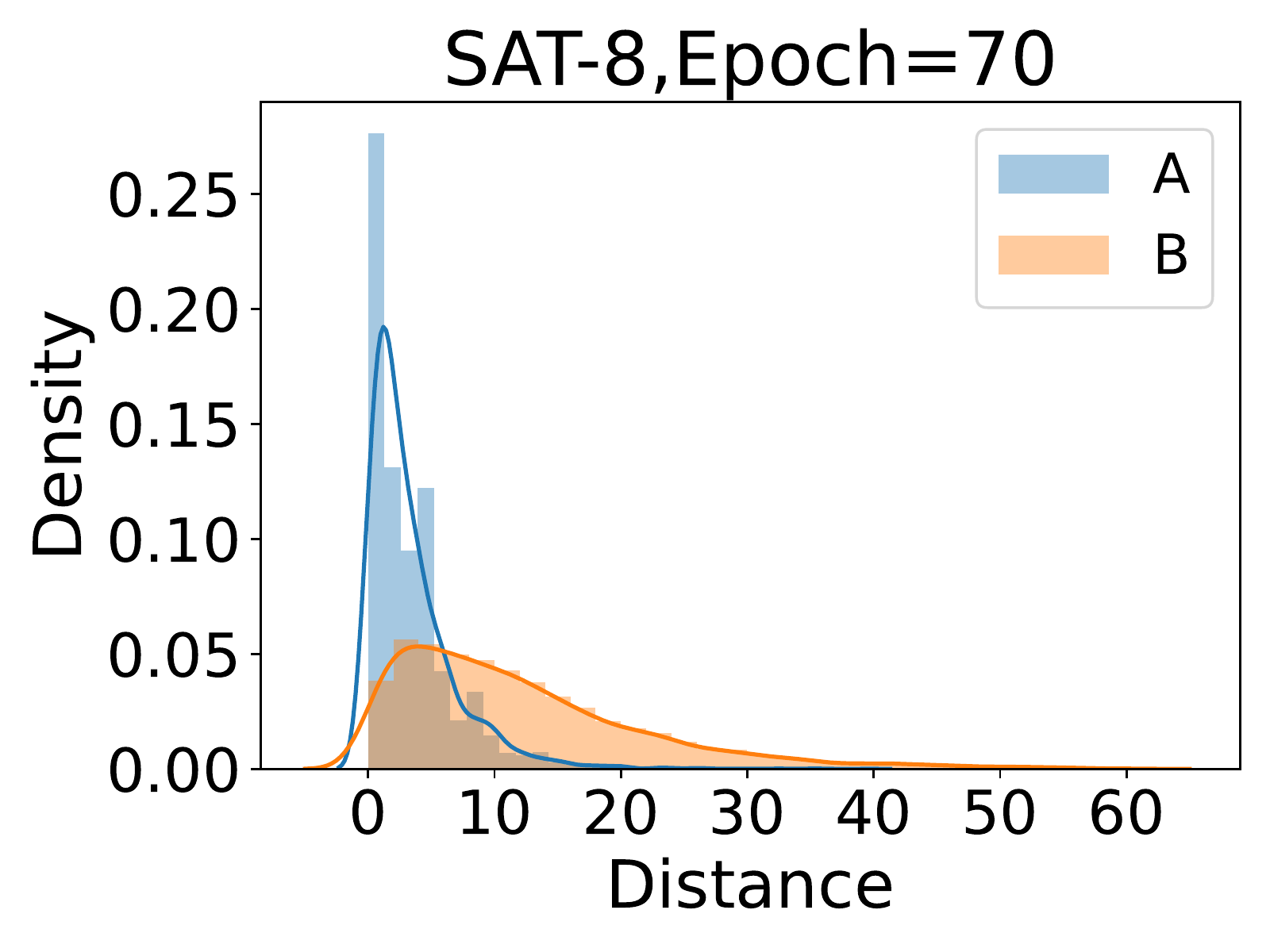}

\end{minipage}
}
\subfigure[(d)]{
\begin{minipage}[t]{0.22\linewidth}
\centering
\includegraphics[width=1.5in,trim=0 0 0 0]{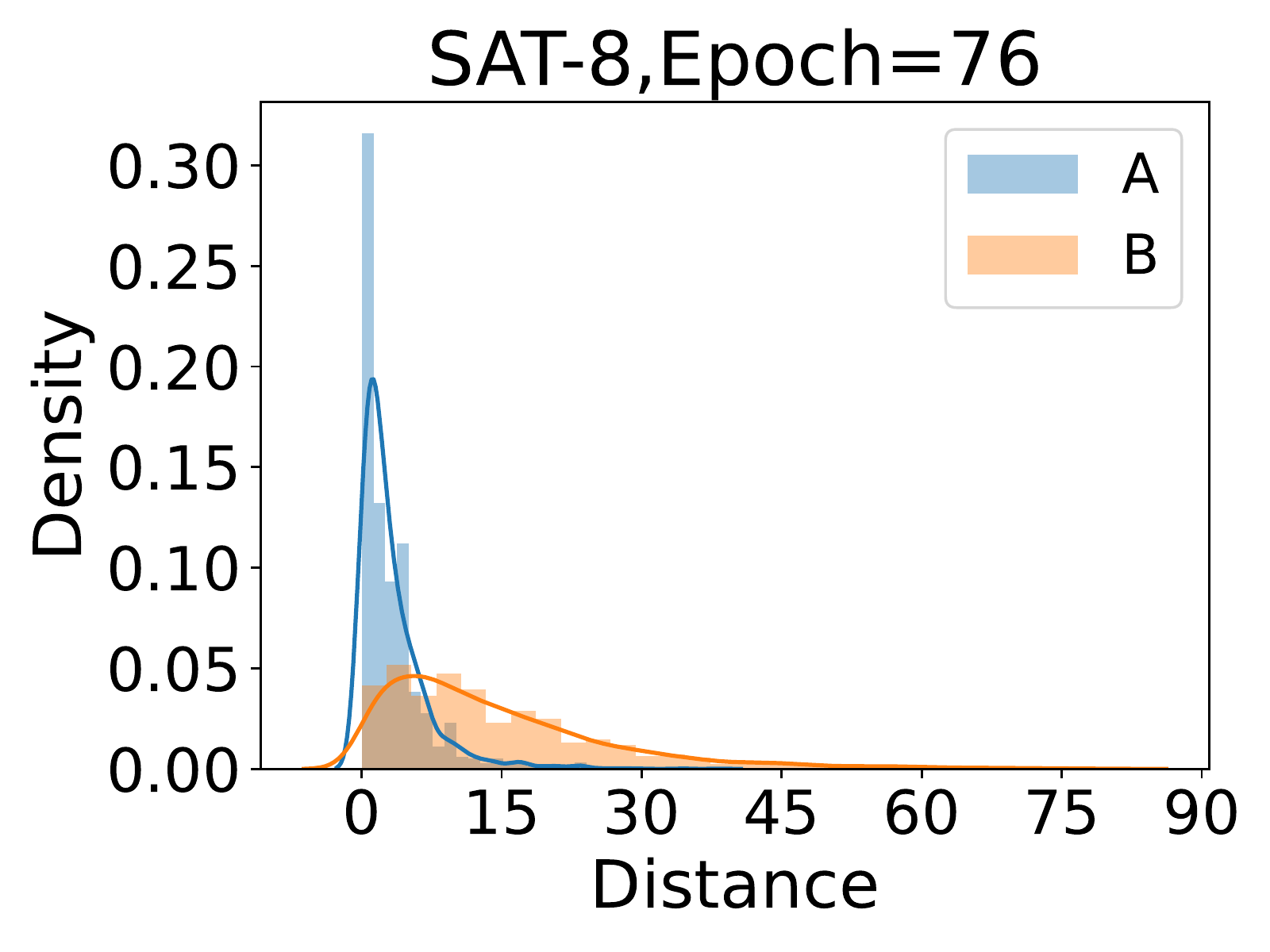}

\end{minipage}
}
\caption{The distribution of examples' distance from different decision boundaries. $\bf{A}$ represents correctly classified examples by the model, while $\bf{B}$ represents misclassified examples. Here we pick the decision boundaries learned by SAT-4 and SAT-8 at different epochs.}
\label{fig3}
\end{figure*}

\subsection{Finer-grained adversarial examples}
In this subsection, we elaborate on the generation of our finer-grained adversarial examples. Intuitively, we can learn $worse$ examples with larger perturbation budgets than $\epsilon_c$ to defend against attacks with that $\epsilon_c$. However, previous adversarial training mostly treats all examples equally to generate adversarial examples, which are also called rough adversarial examples here. It results in: 1) increasing the perturbation budget is accompanied by a significant decrease in natural accuracy; 2) large perturbation budgets are extremely unreasonable for misclassified examples or even correctly classified examples near the decision boundary; and 3) excessive perturbation budgets are not necessarily beneficial for robustness improvement.
\paragraph{\bf{Idea of attack strategy}}Instead of using a uniform $\epsilon$ for all examples, our new attack strategy aims to design finer-grained attack parameters $\epsilon_i$ for them. As we know, the network parameter $\bm \theta$ is updated each epoch during the training process to obtain the current model $f_{\bm \theta_{new}}$. Natural training examples can be divided into two subsets with respect to $f_{\bm \theta_{new}}$, with one subset of correctly classified examples $\mathbb D^+=\{\bm x_i:i \in [N],f_{\bm \theta_{new}}(\bm x_i)=y_i\}$, and one subset of misclassified examples $\mathbb D^-=\{\bm x_i:i \in [N],f_{\bm \theta_{new}}(\bm x_i)=y_i\}$. As shown in Fig. \ref{fig3}, there is a fundamental difference between correctly classified and misclassified examples of a model, with the latter being inherently difficult to learn and inappropriate for further adversarial attack, i.e., $\epsilon_i=0$. Actually, the margin distribution of correctly classified examples in Fig. \ref{fig3} also shows significant differences between these examples, so we calculate margins for the examples correctly classified by the model $f_{\bm \theta_{new}}$ for each epoch, and thus determine their attack parameters. Apparently, our design is more reasonable, which takes into account the multiple characteristics of the examples. In addition, for each epoch, we select a new subset $\mathbb D^+$ and calculate the corresponding margin based on the current network $f_{\bm \theta_{new}}$. Because both the subset and the margin are changing, this new attack scheme belongs to a dynamic adjustment and demonstrates great flexibility.
\paragraph{\bf{Measure of example's margin}}A key point of our attack strategy is the calculation of margin. Since calculating the exact margin in the input space is intractable, we have to search for an approximate solution. To be more precise, we search for a small perturbation $\bm \delta$ such that $f_{\bm \theta}(\bm x_i + \bm \delta)\neq f_{\bm \theta}(\bm x_i)$. Ideally, we then have $m(\bm x_i,\bm v)\approx\| \bm \delta\|_\infty$, although $\bm x_i+\bm \delta$ is not necessarily an element of $\mathbb D'$. We will obtain the margin estimate with the help of DeepFool \cite{a4}, an iterative adversarial attack which stops as soon as a perturbation $\bm \delta$ has been found with $f_{\bm \theta}(\bm x_i + \bm \delta)\neq f_{\bm \theta}(\bm x_i)$. In every iteration step of DeepFool, it linearizes the decision boundary around an example to gradually push the example over the closest boundary for minimal perturbation. This perturbation $\bm \delta_i$ can be written as:

\begin{align}\label{eq10}
\hspace{-2mm}
\bm \delta_i := &\frac{|\bm p_{\hat l}(\bm x_i) - \bm p_{y_i}(\bm x_i)|}{\|\bigtriangledown \bm p_{\hat l}(\bm x_i) - \bigtriangledown \bm p_{y_i}(\bm x_i)\|_1}* (\bigtriangledown \bm p_{\hat l}(\bm x_i) - \bigtriangledown \bm p_{y_i}(\bm x_i))
\end{align}
with index:
\begin{small}
\begin{align}\label{eq11}
\hat l = \hat l(\bm x_i):= \underset{k \neq y_i}{\operatorname {arg\,min}}\frac{|\bm p_{\hat l}(\bm x_i) - \bm p_{y_i}(\bm x_i)|}{\|\bigtriangledown \bm p_{\hat l}(\bm x_i) - \bigtriangledown \bm p_{y_i}(\bm x_i)\|_1}
\end{align}
\end{small}Let $I$ denote the stopping index of this iterative scheme for $\bm x_i$, i.e., $f_{\bm \theta}(\bm x_i^I) \neq f_{\bm \theta}(\bm x_i)$. The desired adversarial perturbation is then defined as $\bm \delta=\sum_{i=0}^{I-1}\bm \delta_i$. Note that we only use successful adversarial perturbations for the margin approximation. Thus, if the DeepFool attack is not able to find a small adversarial perturbation for a given image, we regard its semantic features as more obvious, and the vulnerability can be ignored without considering its margin. We have $m=\| \bm \delta\|_\infty$ now.
\paragraph{\bf{Implementation of attack strategy}} We will use a strategy model to develop our attack strategy, an example-dependent strategy to set the parameter $\epsilon_i$ when performing an adversarial attack, as shown in the Algorithm \ref{alg1}. Since most previous adversarial training work has used SAT-8 ($\epsilon={8/255}$) as a baseline and followed its attack setup, here we use a well-trained SAT-8 as a strategy model and consider the attack scheme to be migratory and applicable to other defense methods. The margin distribution of correctly classified examples of this strategy model is shown in Fig. \ref{fig3}(c). In fact, it is feasible for us to rank the margin values of examples in ascending order and to rate them into three grades according to a certain proportion. The examples at each grade are recorded as $\mathbb D_A, \mathbb D_B, \mathbb D_C$. The example set closest to the decision boundary, $\mathbb D_A$, is the key to determine the degree of destruction of natural accuracy and is our key concern, accounting for 40\%; $\mathbb D_B$, the example set second closest to the decision boundary, the average margin is slightly greater than ${8}/{255}$; $\mathbb D_C$, the example far from the decision boundary, basically completely avoid the cross-over between the adversarial examples and the original natural examples, so $\mathbb D_B$ and $\mathbb D_C$ can well improve the robustness of the model, and the two account for 30\% respectively. The three grades are practically meaningful and sufficiently representative of the properties of correctly classified examples. $\mathbb D_A$, in particular, aims to ensure natural accuracy at the expense of robustness; $\mathbb D_B$ has the greatest trade-off advantage; and $\mathbb D_C$ can further exploit the potential robustness without requiring an excessive range of budgets. As a result, we set $\epsilon_A$, $\epsilon_B$, and $\epsilon_c$ to the values shown in the Algorithm \ref{alg1}. Finally, the FAEs are solved by:

\begin{align}\label{eq0}
\bm {{\tilde x}_i}' = \underset{\bm {x_i}' \in \mathbb{B}(\bm x_i,\epsilon_{i})}{\operatorname {arg\,max}}CE( {{\bm p}(\bm {x_i}'),y_i} )
\end{align}
Based on the two proposed loss terms and the FAEs, we can state our final objective function for MMAT:

\begin{align}\label{eq7}
\hspace{-3mm}
\min_\theta \frac{1}{N} \left(\sum_{i\in \mathbb D} L_1+\sum_{i\in \mathbb D} {L_2}/{\lambda} \right)
\end{align}
where $\lambda$ is a tunable scaling parameter that balances the two parts of the hybrid loss.

The Algorithm \ref{alg2} depicts our Moderate-Margin Adversarial Training. Our attack strategy is based on the margins of examples. The teacher model is used to guide the real-time decision boundary adjustment of our model. As a result, MMAT can also be framed as performing student-teacher learning, particularly self-distillation \cite{c34} to mimic certain properties of the model itself.
\vspace{-1mm}
\begin{table}[b]
\centering
\setlength{\belowcaptionskip}{2.5mm}
\caption{The performance of ResNet-18 is trained using MMAT with different attack settings on CIFAR-10. We pick the checkpoint which has the best robust accuracy on the test set. The difference (Diff.) between best and final accuracy indicates degradation in performance during training.}
\label{tbl1}
\scalebox{0.9}{
\begin{tabular}{cccccccc}
\toprule
     &       &       & RA    &       &       & NA    &       \\ \cmidrule{3-8}
     &       & Best  & Final & Diff. & Best  & Final & Diff. \\ \midrule
 & 2     & 55.58 & 55.14 & 0.44  & 85.32 & 85.84 & -0.52 \\
$Z_1$     & 3     & 53.77 & 53.31 & 0.46  & 86.36 & 86.54 & -0.18 \\
     & 4     & 52.13 & 51.49 & 0.64  & 87.71 & 87.80 & -0.09 \\ \midrule
 & 3/255 & 55.16 & 54.57 & 0.59  & 85.15 & 85.54 & -0.39 \\
$\epsilon_A$     & 5/255 & 55.58 & 55.14 & 0.44  & 85.32 & 85.84 & -0.52 \\
     & 7/255 & 55.69 & 55.19 & 0.50  & 84.77 & 85.25 & -0.48 \\ \bottomrule
\end{tabular}}
\end{table}
\begin{table}[t]
\centering
\setlength{\belowcaptionskip}{2.5mm}
\caption{Comparisons with other defense methods under white-box attacks. We pick the checkpoint which has the best robust accuracy on the test set. Note that we still follow the basic experimental setup described in Section \ref{set5.1}. Hyper-parameters and training details of other defense methods are configured as per their original papers: $\epsilon=6$ for TRADES and MART; ResNet18 is adopted as $\rm M_{natural}$ for LBGAT; a well naturally trained $f_{std}$ for HAT.}
\label{tbl2}
\scalebox{0.65}{
\begin{tabular}{llllll}
\toprule
\multirow{2}{*}{Adversary} & \multirow{2}{*}{Method} & \multicolumn{2}{c}{CIFAR-10} & \multicolumn{2}{c}{SVHN} \\
                           &                         & RA(\%)        & NA(\%)       & RA(\%)      & NA(\%)     \\ \toprule
\multirow{6}{*}{FGSM}      & SAT                     & 57.34         & 83.50        & 71.17       & 92.47      \\
                           & TRADES                  & 56.72         & 81.57        & 73.32       & 91.25      \\
                           & MART                    & 59.68         & 81.39        & 72.93       & 91.43      \\
                           & LBGAT                   & 56.75         & 85.28        & 69.55       & 92.75      \\
                           & HAT                     & 61.10         & 85.95        & 76.66       & 92.88      \\
                           & MMAT               & $\bm {59.78}$         & $\bm {85.32}$        & $\bm{72.75}$       & $\bm {93.45}$      \\ \midrule
\multirow{6}{*}{$\rm PGD^{20}$}     & SAT                     & 52.69         & 83.50        & 58.52       & 92.47      \\
                           & TRADES                  & 52.90         & 81.57        & 60.83       & 91.25      \\
                           & MART                    & 55.33         & 81.39        & 60.70       & 91.43      \\
                           & LBGAT                   & 52.10         & 85.28        & 59.84       & 92.75      \\
                           & HAT                     & 52.49         & 85.95        & 61.65       & 92.88      \\
                           & MMAT               & $\bm {55.58}$         & $\bm {85.32}$        & $\bm{ 62.20}$       &$\bm{ 93.45}$      \\ \midrule
\multirow{6}{*}{$\rm CW_\infty$}       & SAT                     & 49.23         & 83.50        & 54.64       & 92.47      \\
                           & TRADES                  & 49.41         & 81.57        & 56.16       & 91.25      \\
                           & MART                    & 49.91         & 81.39        & 53.40       & 91.43      \\
                           & LBGAT                   & 48.81         & 85.28        & 54.92       & 92.75      \\
                           & HAT                     & 48.73         & 85.95        & 55.27       & 92.88      \\
                           & MMAT               & $\bm {50.33}$         & $\bm {85.32}$        & $\bm {55.35}$       & $\bm {93.45}$      \\ \midrule
\multirow{6}{*}{AA}        & SAT                     & 47.67         & 83.50        & 50.12       & 92.47      \\
                           & TRADES                  & 48.33         & 81.57        & 53.44       & 91.25      \\
                           & MART                    & 48.03         & 81.39        & 49.83       & 91.43      \\
                           & LBGAT                   & 47.29         & 85.28        & 52.02       & 92.75      \\
                           & HAT                     & 47.28         & 85.95        & 51.06       & 92.88      \\
                           & MMAT               & $\bm {48.49}$         & $\bm {85.32}$        & $\bm {52.13}$       & $\bm {93.45}$      \\ \bottomrule
\end{tabular}}
\end{table}
\section{Experiment}
\subsection{Experimental
 setup}\label{set5.1}
\paragraph{\bf{Training methods}}We consider neural networks trained via Natural Training (NT), Standard Adversarial Training (SAT) \cite{c3}, TRADES \cite{c11}, MART \cite{c1}, LBGAT \cite{c13}, and HAT \cite{c14}. The trade-off parameter, $\beta$, is present in TRADES, MART, LBGAT, and HAT, and a higher $\beta$ indicates that the robust accuracy is given more weight.
\vspace{-1mm}
\paragraph{\bf{White-box attacks}}We evaluate robustness with four types of attacks, \textit{i.e.}, FGSM, $\rm PGD^{20}$, $\rm CW_\infty$, and Auto-Attack (AA), on two datasets. All attacks are constrained by the same perturbation budget, \textit{i.e.}, $\epsilon={8}/{255}$. Note that the $\rm CW_\infty$ attack denotes using CW-loss within the PGD framework here. The goal of Auto-Attack is to reliably evaluate model robustness with an ensemble of diverse strong attack methods. We use the open-source code from \cite{c7} to test our models. Besides, for PGD, the training attack is $\rm PGD^{10}$ with random start and step size $\alpha={\epsilon}/{4}$, while the test attack is $\rm PGD^{20}$ with random start, perturbation budget $\epsilon={8/255}$ and step size $\alpha={\epsilon/10}$.
\vspace{-2mm}
\paragraph{\bf{Black-box attacks}}We train ResNet-18 on CIFAR-10 and PreActResNet-18 on SVHN with different adversarial training methods. Then we use the well-trained model as the source model or the target model. Adversarial examples are generated from the source model to attack the target model. On both datasets, the $\rm PGD^{20}$(black-box) is applied to attack various defense models. And the setup follows that specified in the white-box attack.

\paragraph{\bf{Training details}}To evaluate the performance of different models, we conduct extensive experiments on ResNet-18 \cite{b1} using CIFAR-10 \cite{b2} and PreActResNet-18 using SVHN \cite{b3}. For CIFAR-10, following \cite{c1}, the batch size is 128 and the number of training epochs is 100. All models are trained using the SGD optimizer with Nesterov momentum 0.9 \cite{b4}, weight decay 0.0035 and an initial learning rate of 0.01, which is divided by 10 at the 75-th and 90-th epochs. Data augmentation is performed. When performing data augmentation, we randomly crop the image to $32 \times 32$ with 4 pixels of padding, then perform random horizontal flips. For SVHN, following \cite{c13}, the batch size is still 128. We use the SGD optimizer with Nesterov momentum 0.9 and weight decay 0.0005. We further employ cyclic learning rates \cite{b5} with cosine annealing and a maximum learning rate of 0.05 for SVHN. Here we only run 15 epochs.
\begin{figure}[t]
\setlength{\abovecaptionskip}{0cm}
\setlength{\belowcaptionskip}{-0.2cm}
\centering
\subfigure[(a)]{
\begin{minipage}[t]{0.4\linewidth}
\centering
\includegraphics[width=1.3in,height=2.6cm,trim=0 0 0 0]{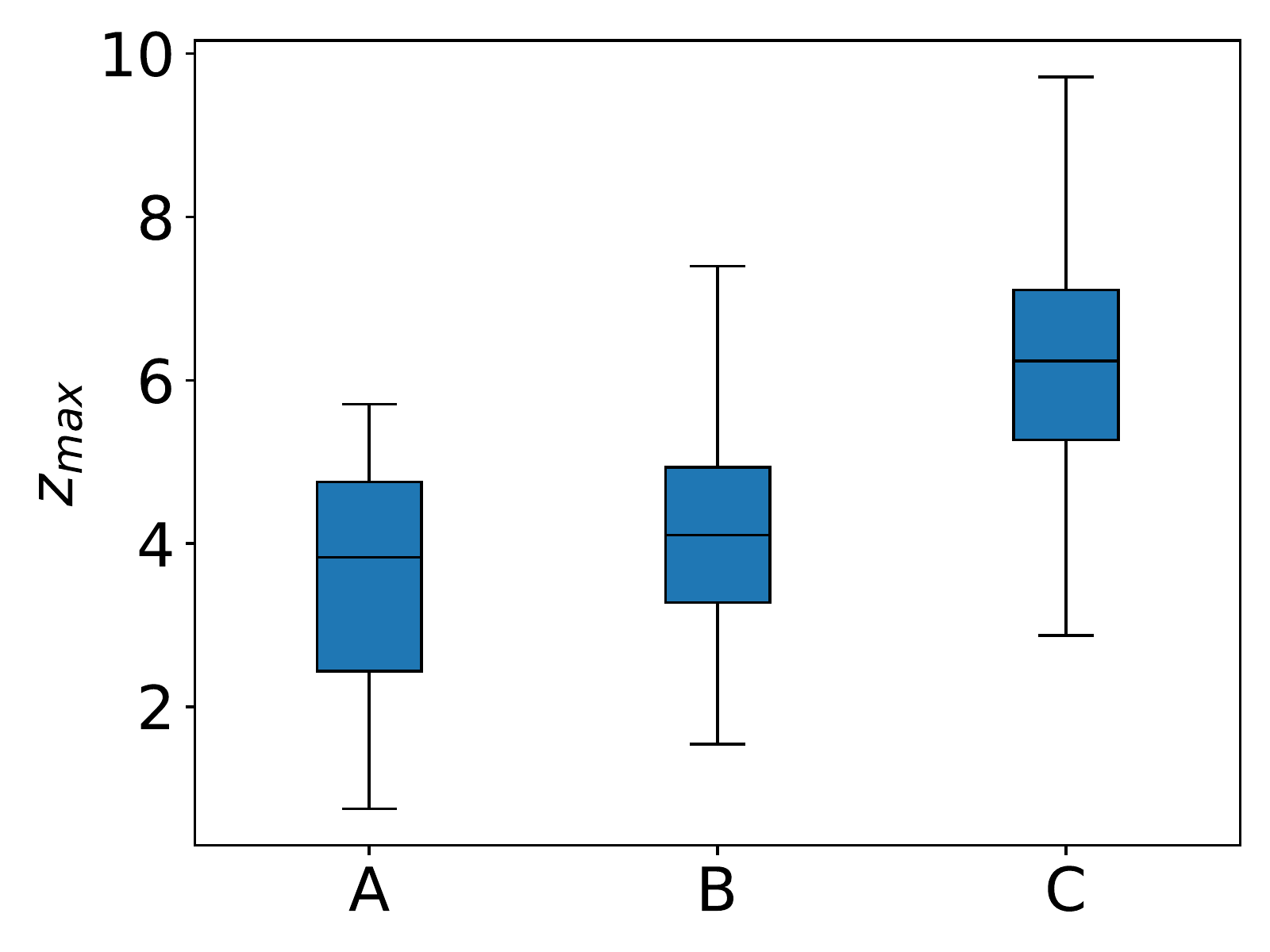}
\end{minipage}
}
\subfigure[(b)]{
\begin{minipage}[t]{0.4\linewidth}
\centering
\includegraphics[width=1.4in,height=2.6cm,trim=0 0 0 0]{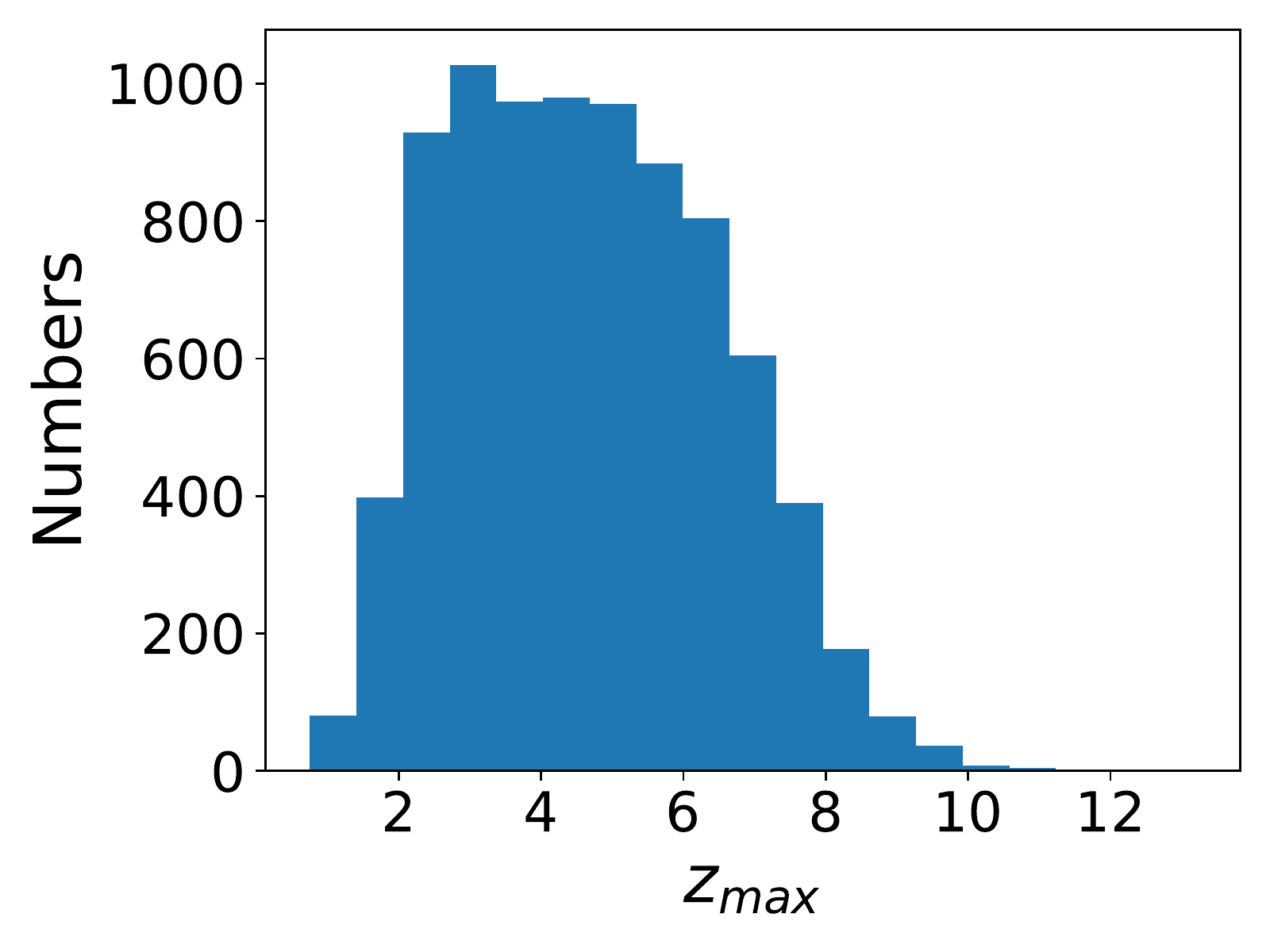}
\end{minipage}
}
\caption{The analysis of ${\bm z}_{max}$ of correctly classified examples by the well-trained strategy network. (a): The horizontal axis represents the grades based on the margins of these examples. Examples with higher grades correspond to larger ${\bm z}_{max}$. (b): The distribution of ${\bm z}_{max}$. Here we sample 10000 examples from the training examples.}
\label{fig4}
\end{figure}

\begin{figure*}[t]
\setlength{\abovecaptionskip}{0.1cm}
\setlength{\belowcaptionskip}{-0.4cm}
\centering
\subfigure[(a)]{
\begin{minipage}[t]{0.2\linewidth}
\centering
\includegraphics[width=1.5in]{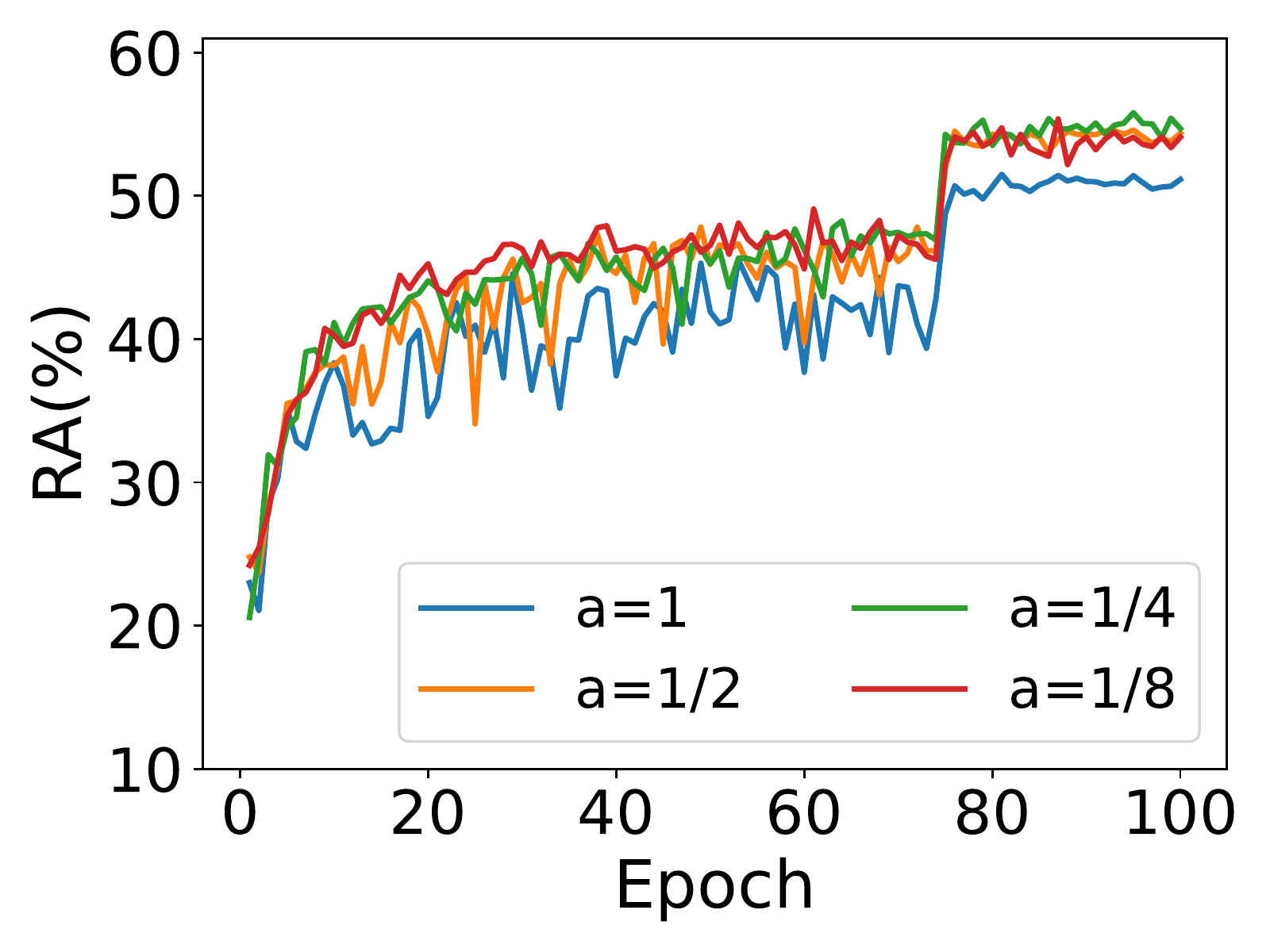}
\end{minipage}
}
\subfigure[(b)]{
\begin{minipage}[t]{0.2\linewidth}
\centering
\includegraphics[width=1.5in]{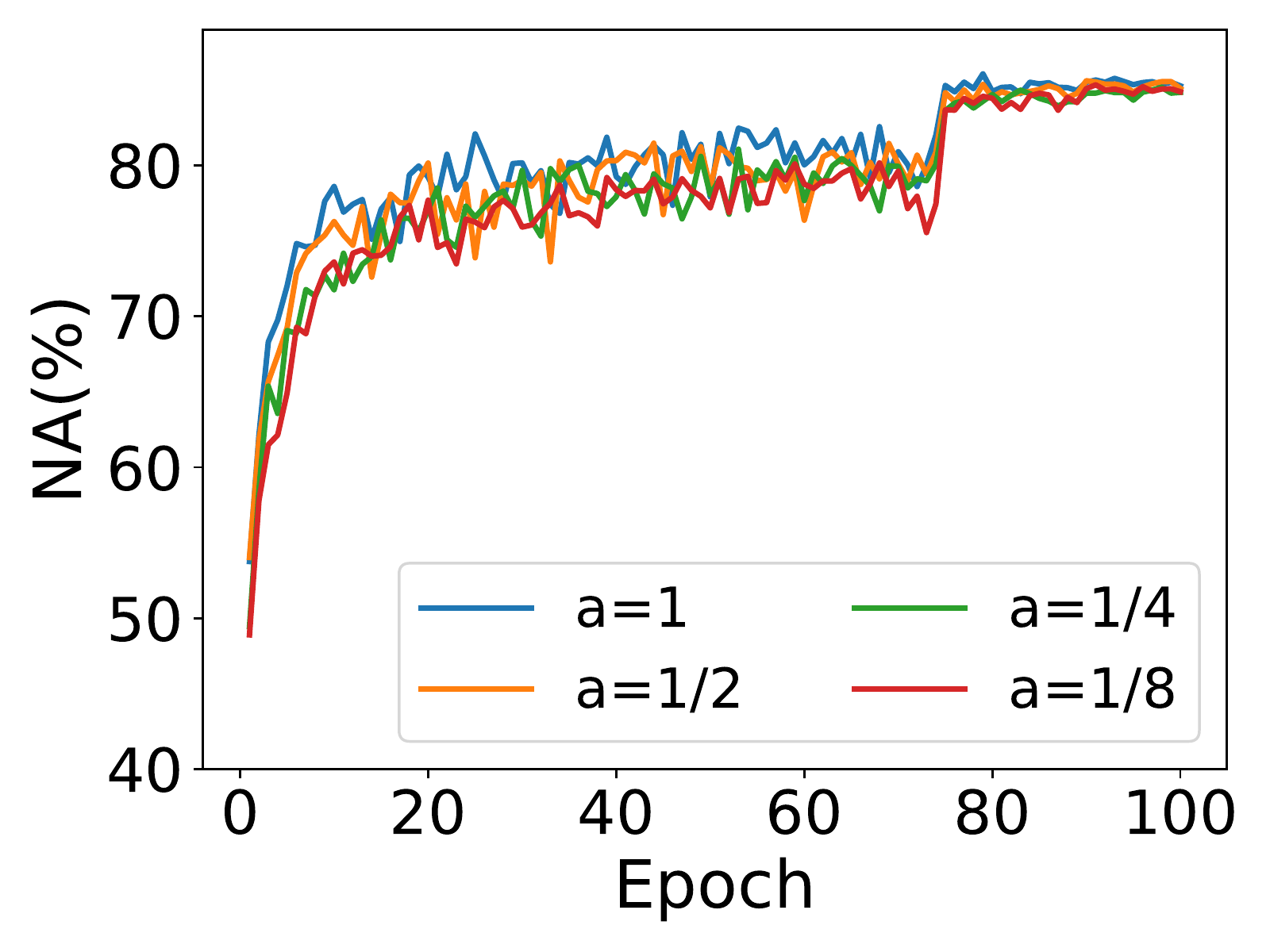}
\end{minipage}
}
\subfigure[(c)]{
\begin{minipage}[t]{0.2\linewidth}
\centering
\includegraphics[width=1.5in]{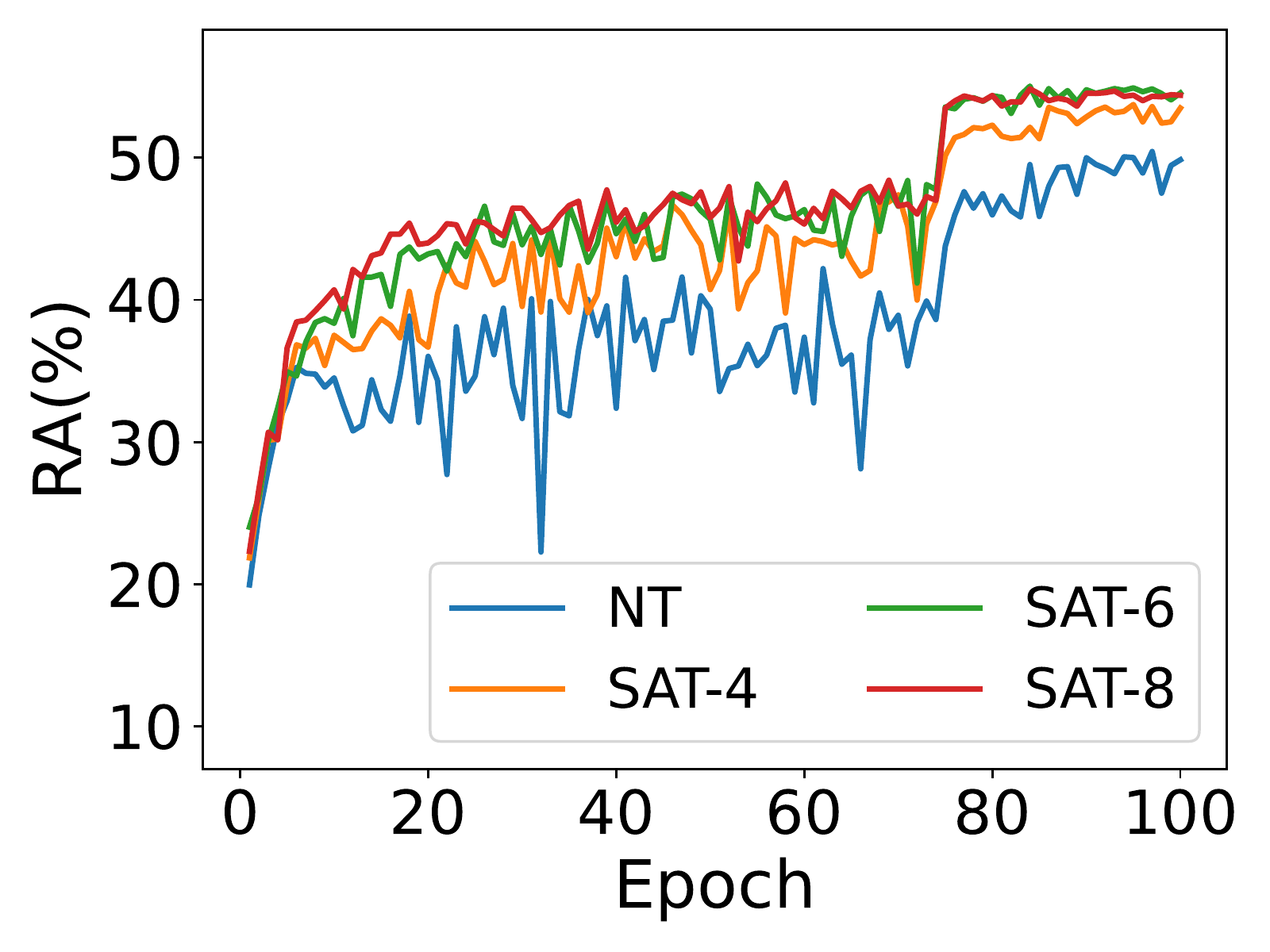}

\end{minipage}
}
\subfigure[(d)]{
\begin{minipage}[t]{0.2\linewidth}
\centering
\includegraphics[width=1.5in]{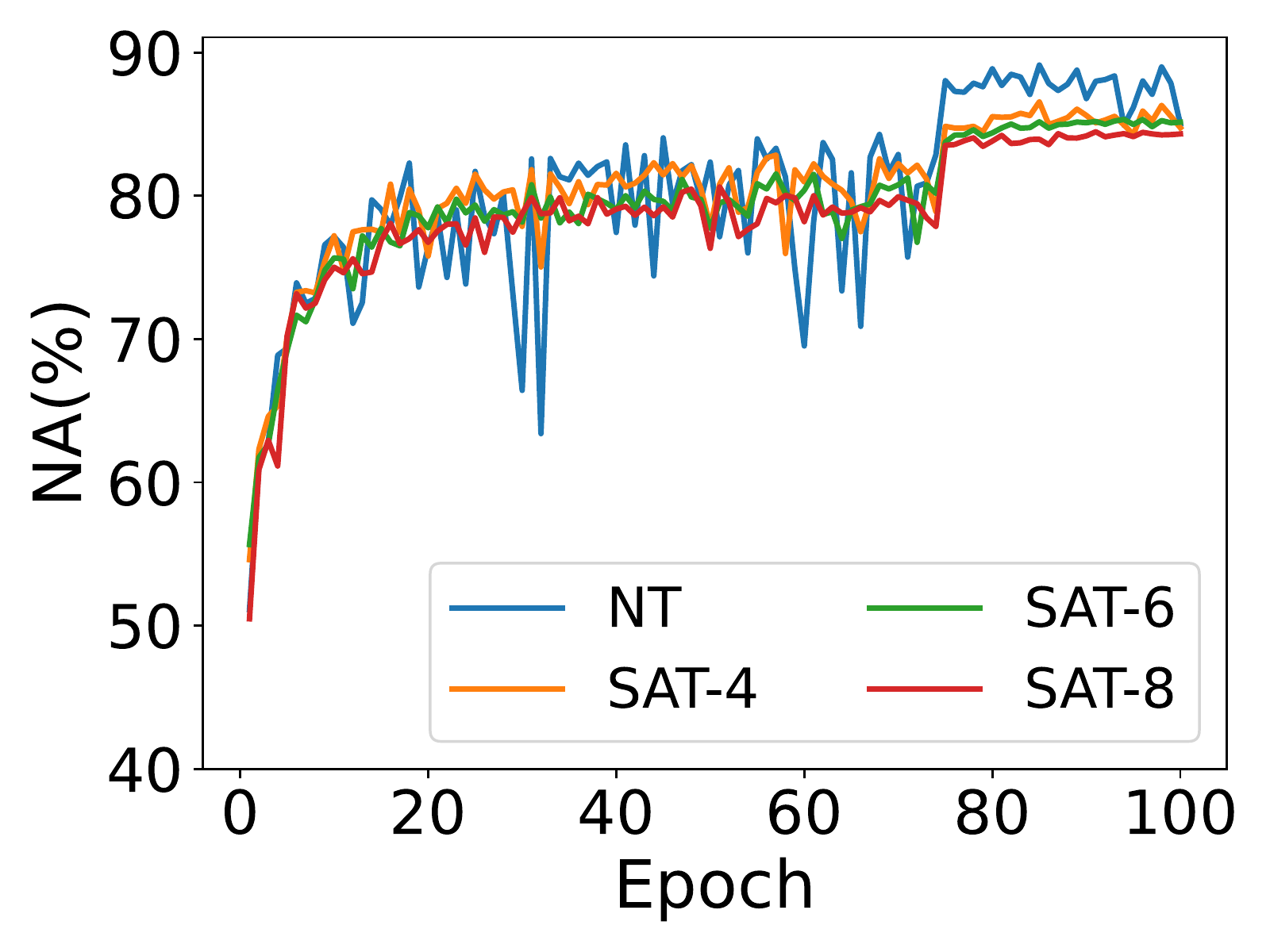}

\end{minipage}
}
\caption{The guiding role of teacher models. (a)-(b): Our models are trained with different scaling parameters $a$ (1/$\lambda$) under the supervision of SAT-6. (c)-(d): Our models are trained under the supervision of four teacher models, each representing a different level of NA and obtained through various training methods.}
\label{fig5}
\end{figure*}
\subsection{Understanding the proposed MMAT}
In this section, we investigate MMAT from three different perspectives: 1) the perturbation budgets of FAEs; 2) the guiding role of teacher models; and 3) replacing components of the BCE term.

\paragraph{\bf{Perturbation budget of FAEs}}Since adversarial training is time-consuming, taking 3-30 times longer to form a robust network than a non-robust equivalent \cite{c35}, \cite{c36}, we prefer to use a metric without extra computation to determine the grade of examples instead of the margin. As shown in Fig. \ref{fig4}(a), $z_{max}$(the largest element of a network's logit output after forward- propagation) of examples meet this requirement. In Fig. \ref{fig4}(b), the distribution of $z_{max}$ implies the hierarchical nature of these examples; we simply use $Z_1=2$, $Z_2=6$ to divide these examples into three grades, i.e., $\mathbb D_A = \{\bm x_i|\bm z_{max}(\bm x_i)\leq2\}$, $\mathbb D_B = \{\bm x_i|\bm z_{max}(\bm x_i)\leq6\}$, $\mathbb D_C = \{\bm x_i|\bm z_{max}(\bm x_i)>6\}$. And, according to Algorithm \ref{alg1}, $\epsilon_A={5}/{255}$, we have $\epsilon_B={10}/{255}$, $\epsilon_A={15}/{255}$, which means that $\epsilon_i$ of each example $\bm x_i$ can be set. Furthermore, we investigate the effects of $\mathbb D_A$ under various $Z_1$ and $\epsilon_A$, with the results summarized in Table \ref{tbl1}. Apparently, larger $Z_1$ or smaller $\epsilon_A$ appears to favor natural accuracy over robustness. The results demonstrate the powerful regulation capability of MMAT in terms of the RA-NA trade-off. We also discovered that MMAT can enable larger perturbation budget $\epsilon_i$ by alleviating the cross-over mixture problem \cite{c18}.

\paragraph{\bf{Role of teacher models}}We first investigate the parameter $a$ (1/$\lambda$) in the MMAT objective function defined in \eqref{eq7} which controls the strength of the regularization. A larger $a$ indicates that the MSE term plays a more significant role, and the teacher model has a more obvious influence on the course of the decision boundary. As illustrated in Fig. \ref{fig5}(a) and (b), as $a$ increases, the RA in the training process tends to decrease, especially when $a=1$, while NA tends to increase. For $a=1/4$ and $a=1/8$, the model has higher RA and NA, which confirms that the fine-tuning effect of the teacher model can maintain high NA without compromising the robustness of the student model. Next, fixing $a=1/4$, we further explore the effect of different teacher models. As the teacher model becomes more robust with lower NA, the student model also shows this trend. Thus, we confirm that fine-tuning of a suitable teacher model facilitates the desirable course of decision boundary to achieve a good trade-off.

\begin{figure*}[t]
\setlength{\abovecaptionskip}{0.1cm}
\setlength{\belowcaptionskip}{-0.3cm}
\centering
\subfigure[(a)]{
\begin{minipage}[t]{0.2\linewidth}
\centering
\includegraphics[width=1.5in]{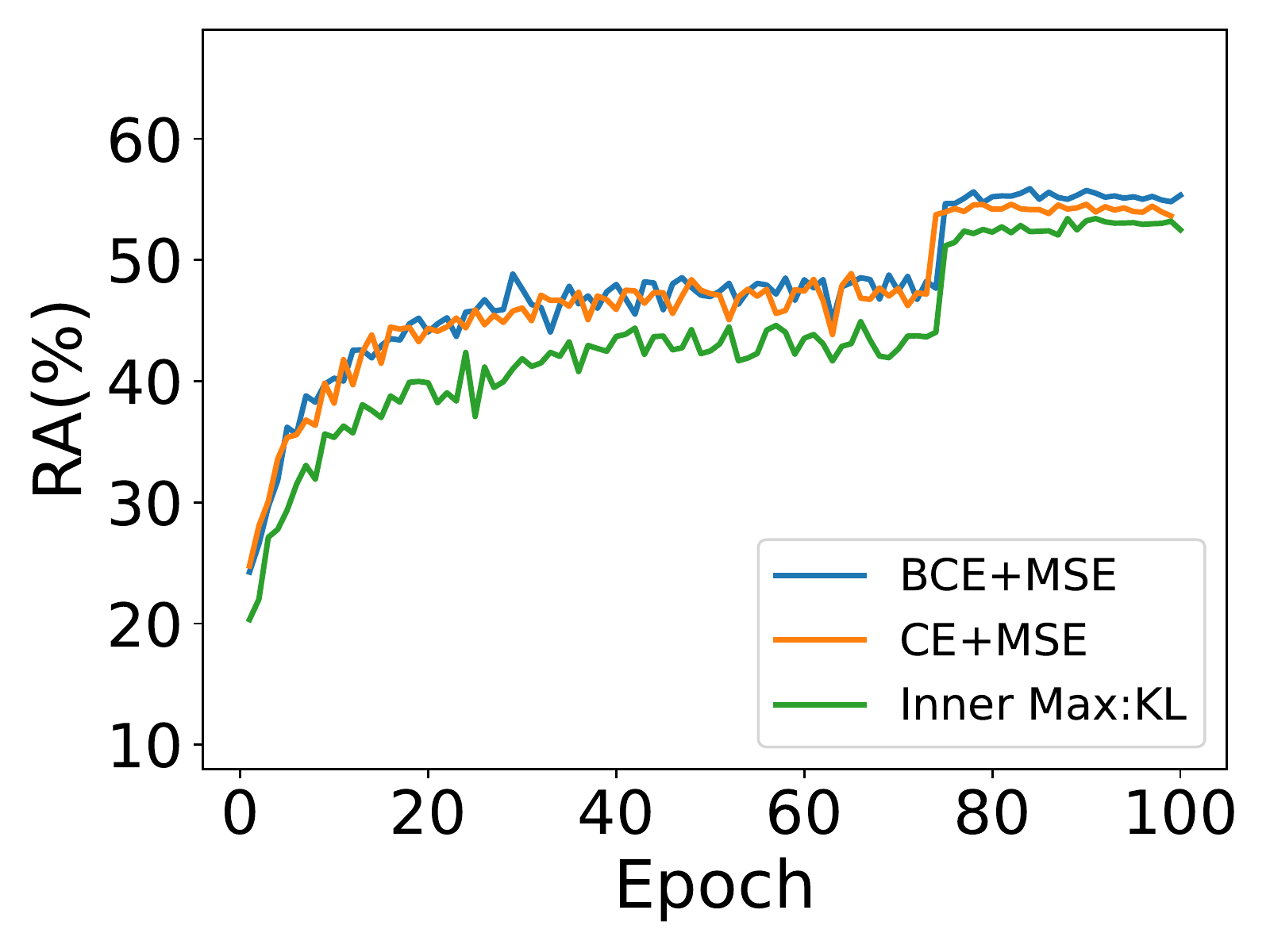}
\end{minipage}
}
\subfigure[(b)]{
\begin{minipage}[t]{0.2\linewidth}
\centering
\includegraphics[width=1.5in]{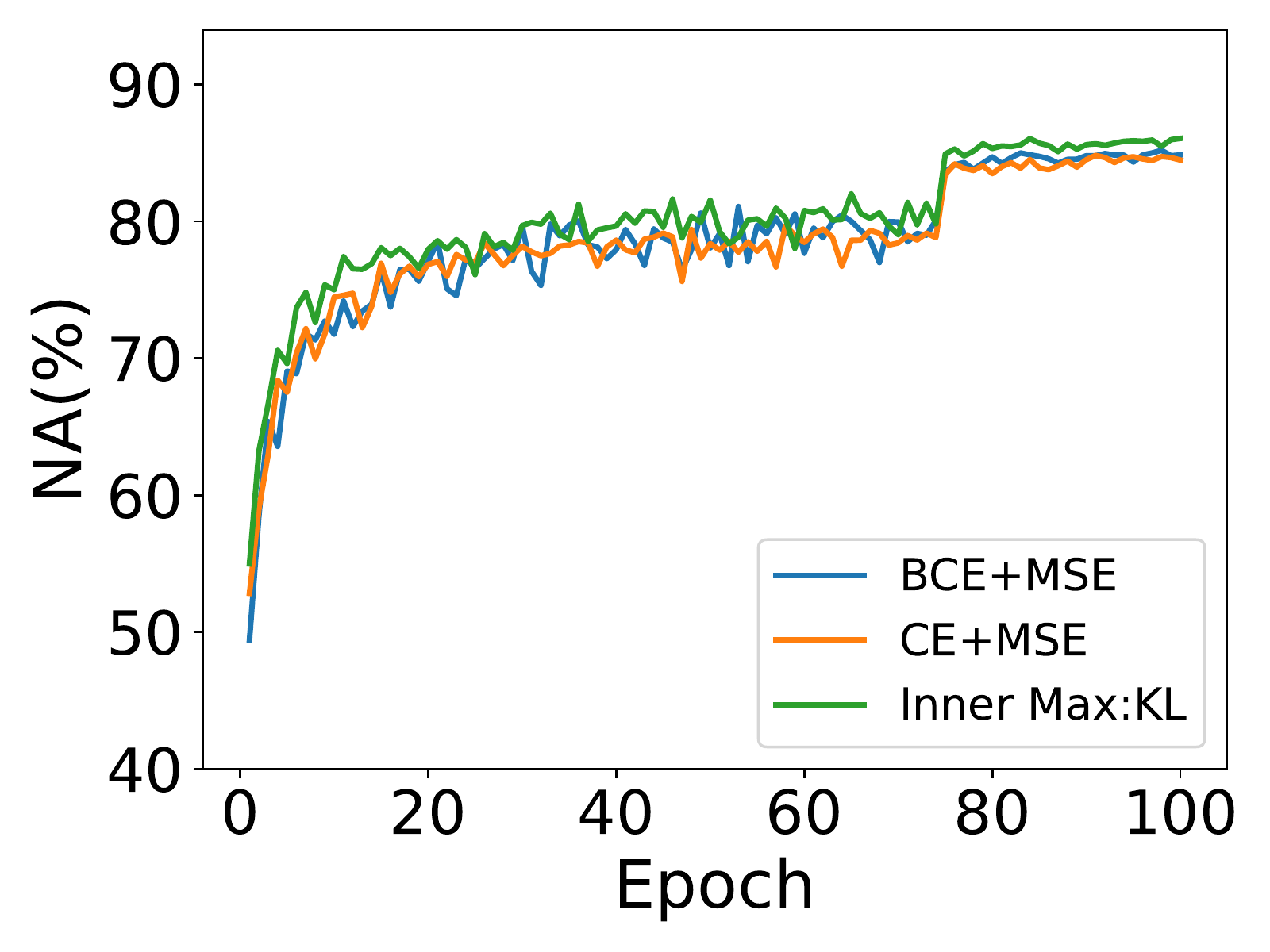}
\end{minipage}
}
\subfigure[(c)]{
\begin{minipage}[t]{0.2\linewidth}
\centering
\includegraphics[width=1.5in]{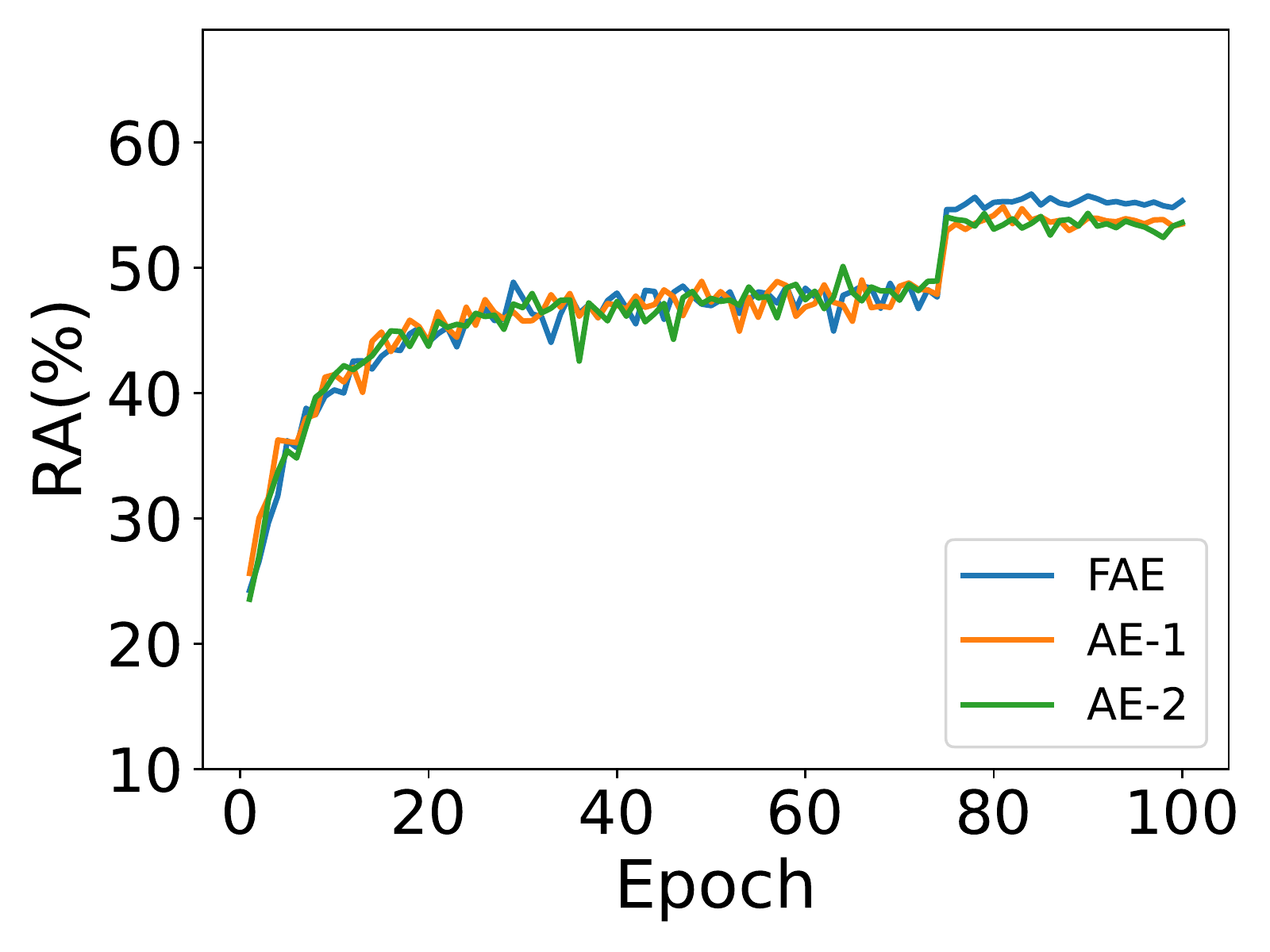}

\end{minipage}
}
\subfigure[(d)]{
\begin{minipage}[t]{0.2\linewidth}
\centering
\includegraphics[width=1.5in]{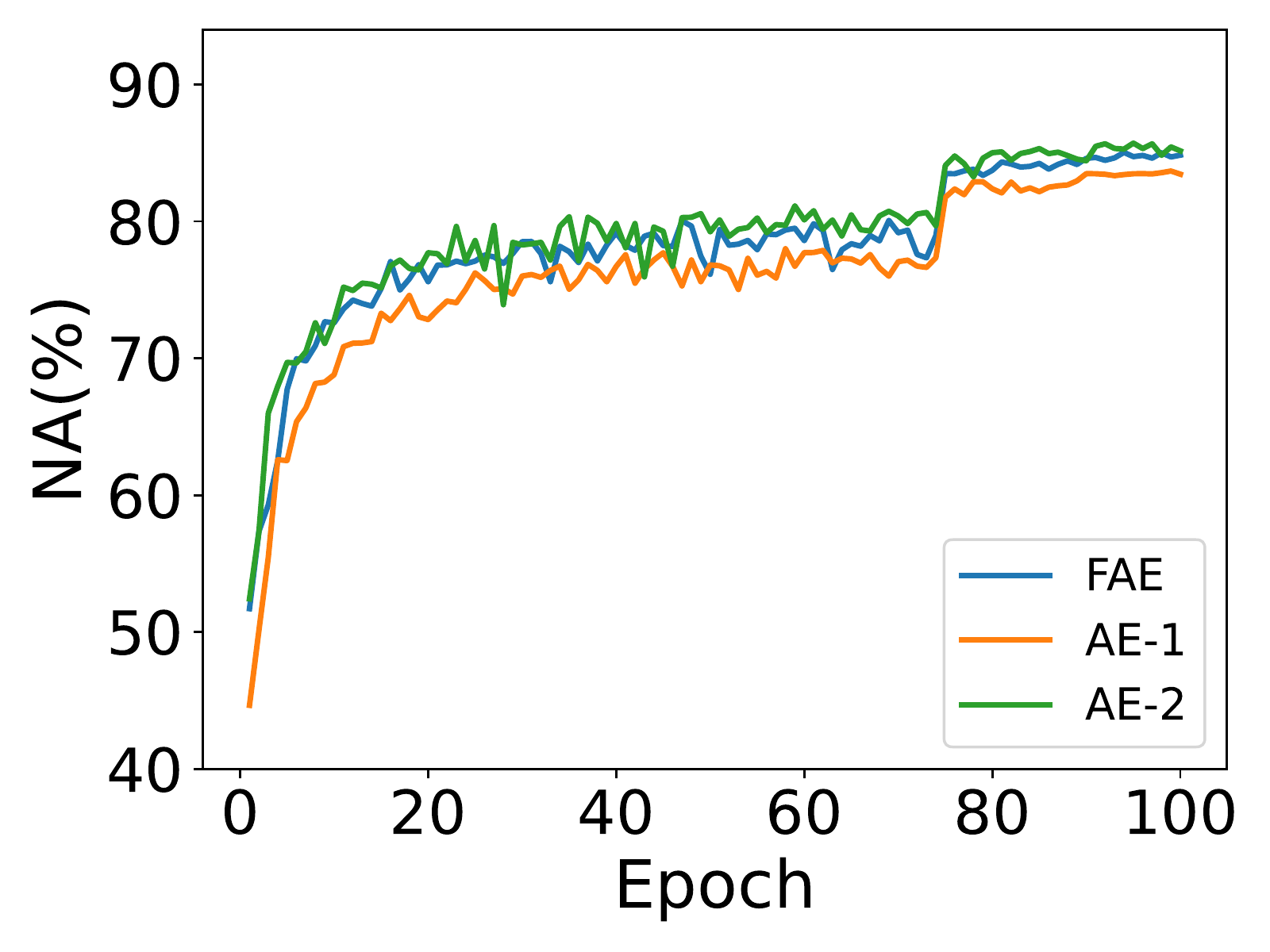}

\end{minipage}
}
\caption{We replace components of the BCE term and compare the performance of the models trained with MMAT variants. In each plot, the solid blue line represents the original MMAT method. (a)-(b): We replace BCE loss with CE loss. In the adversarial min-max framework, we use KL instead of CE for inner maximization. (c)-(d): We substitute other attack strategies for our FAEs. FAE represents our original MMAT. AE-1 is an attack strategy that no longer distinguishes whether an example is correctly classified or not, but directly ranks all examples. AE-2 is an attack that adheres to the uniform perturbation budget of SAT and only attacks correctly classified examples.}
\label{fig6}
\end{figure*}
\paragraph{\bf{Replacing components of $L_1$ term}}Recall that loss term $L_1$ is defined by BCE. As we show in Fig. \ref{fig6}(a) and (b), learning with CE instead of our BCE suffers from insufficient learning with lower robustness throughout the entire training process. Similarly, when replacing CE with KL in the inner maximization of the adversarial min-max framework, the robustness has been weak throughout the training process. Fig. \ref{fig6}(c) and (d) show the contribution of our BCE term with FAEs. Specifically, we generate adversarial examples with three strategies. Both the RA and NA of AE-1 are lower than our original MMAT, suggesting that learning misclassified examples is more necessary than learning their adversarial counterparts. Moreover, AE-2 leads to a robustness degradation, indicating that our FAE associated with characteristics of natural examples better exploits the model's robustness potential.

\begin{table}[t]
\centering
\setlength{\belowcaptionskip}{2.5mm}
\caption{Comparisons with other defense models under black-box $\rm PGD^{20}$ attack on CIFAR-10 and SVHN. Source and target models are the same as those used for white-box attacks. Here, MMAT is used as a source model to attack other defense models, and it is also used as a target model to be attacked by other defense models.}
\scalebox{0.65}{
\begin{tabular}{llllll}
\toprule
{Dataset}                    & Source Model & Target model & RA(\%)     & NA(\%)     &  \\ \midrule
\multirow{10}{*}{CIFAR-10} & MMAT         & SAT          & 63.43 & 83.50 &  \\
                           & MMAT         & TRADES       & 63.26 & 81.57 &  \\
                           & MMAT         & MART         & 63.89 & 81.39 &  \\
                           & MMAT         & LBGAT        & 63.83 & 85.28 &  \\
                           & MMAT         & HAT          & 64.64 & 85.95 &  \\ 
                           & SAT          & MMAT         & 63.95 & 85.32 &  \\
                           & TRADES       & MMAT         & 63.90 & 85.32 &  \\
                           & MART         & MMAT         & 65.64 & 85.32 &  \\
                           & LBGAT        & MMAT         & 64.17 & 85.32 &  \\
                           & HAT          & MMAT         & 64.11 & 85.32 &  \\ \midrule
\multirow{10}{*}{SVHN}     & MMAT         & SAT          & 64.60 & 92.47 &  \\
                           & MMAT         & TRADES       & 66.39 & 91.25 &  \\
                           & MMAT         & MART         & 65.58 & 91.43 &  \\
                           & MMAT         & LBGAT        & 67.20 & 92.75 &  \\
                           & MMAT         & HAT          & 65.66 & 92.88 &  \\ 
                           & SAT          & MMAT         & 65.49 & 93.45 &  \\
                           & TRADES       & MMAT         & 66.88 & 93.45 &  \\
                           & MART         & MMAT         & 68.31 & 93.45 &  \\
                           & LBGAT        & MMAT         & 65.47 & 93.45 &  \\
                           & HAT          & MMAT         & 73.41 & 93.45 &  \\ \bottomrule
\end{tabular}}
\label{tbl3}
\end{table}
\subsection{Comparisons with other defense methods}
In this section, we assess MMAT and other defense methods mentioned in Section \ref{set5.1} against white-box and black-box attacks, as well as perform a deep analysis to confirm the effectiveness of our method.
\paragraph{\bf{Evaluation results under white-box attacks}}Table \ref{tbl2} shows the evaluation results of defense models under white-box attacks on CIFAR-10 and SVHN. Our model achieves high RA and NA across multiple datasets and attack methods. On CIFAR-10, TRADES and MART have relatively low NA but perform well in terms of robustness. LBGAT and HAT have higher NA, but their robustness is even weaker than SAT. It is worth noting that our model's robustness exceeds that of MART, while our NA remains very high, second only to HAT. Because it is easier to classify than CIAFR-10, the NA is generally high, and the difference between them on SVHN is small. Despite this, our method has the best NA. In terms of robustness, TRADES achieves the best results under $\rm CW_\infty$ and AA attacks, LBGAT comes in second only to TRADES, MART loses its original advantage, while HAT and MMAT perform admirably under all four attacks.
\begin{figure*}[t]
\setlength{\abovecaptionskip}{0.2cm}
\setlength{\belowcaptionskip}{-0.3cm}
\centering
\subfigure[(a)SAT]{
\begin{minipage}[t]{0.25\linewidth}
\centering
\includegraphics[width=1.5in,trim=0 0 0 0]{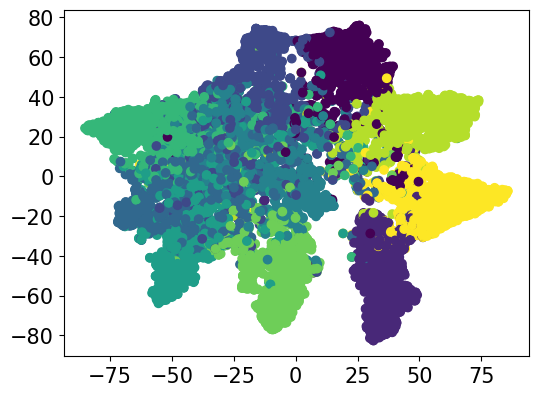}
\end{minipage}
}
\subfigure[(b)TRADES]{
\begin{minipage}[t]{0.25\linewidth}
\centering
\includegraphics[width=1.5in,trim=0 0 0 0]{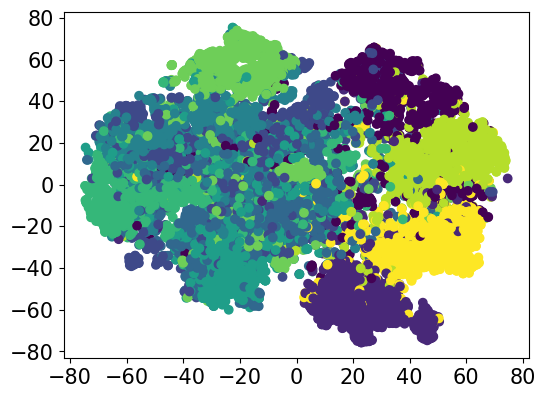}
\end{minipage}
}
\subfigure[(c)MART]{
\begin{minipage}[t]{0.25\linewidth}
\centering
\includegraphics[width=1.5in,trim=0 0 0 0]{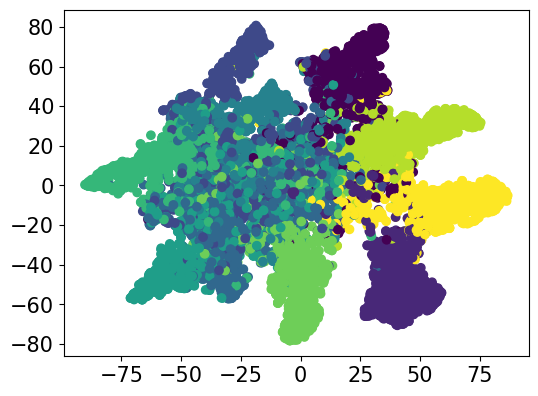}
\end{minipage}
}
\subfigure[(d)LBGAT]{
\begin{minipage}[t]{0.25\linewidth}
\centering
\includegraphics[width=1.5in,trim=0 0 0 0]{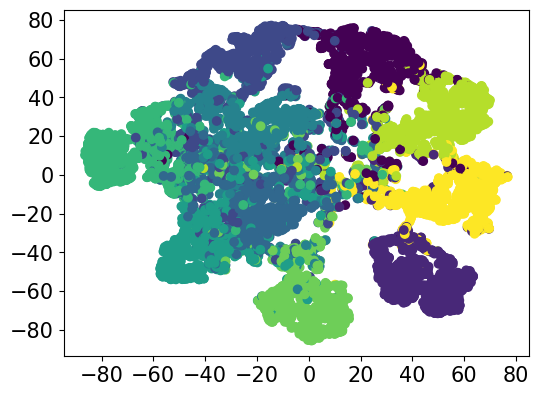}

\end{minipage}
}
\subfigure[(e)HAT]{
\begin{minipage}[t]{0.25\linewidth}
\centering
\includegraphics[width=1.5in,trim=0 0 0 0]{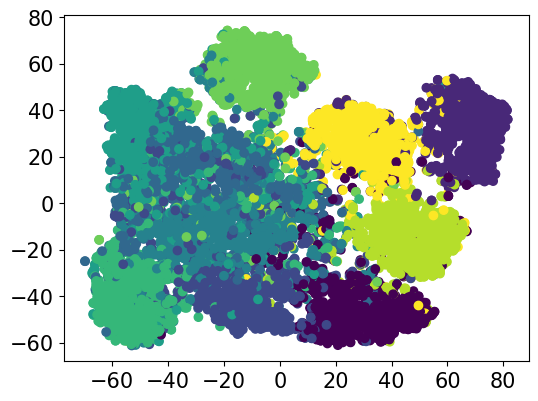}

\end{minipage}
}
\subfigure[(f)MMAT]{
\begin{minipage}[t]{0.25\linewidth}
\centering
\includegraphics[width=1.5in,trim=0 0 0 0]{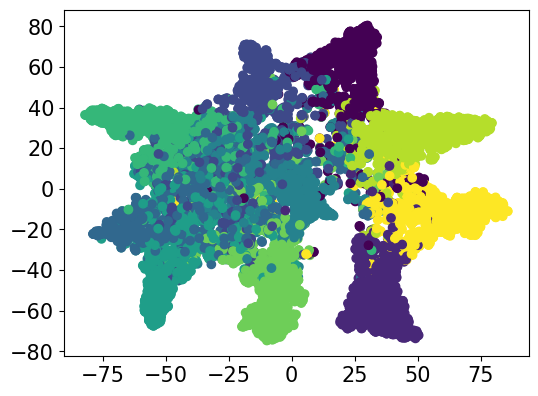}

\end{minipage}
}
\caption{t-SNE results of different methods trained on CIFAR-10. Different colors represent different classes. (a)-(f) represent different training methods, respectively.}
\label{fig7}
\end{figure*}
\paragraph{\bf{Evaluation results under black-box attacks}}The evaluation results of defense models under black-box attacks on CIFAR-10 and SVHN are reported in Table \ref{tbl3}. Compared with the white-box ($\rm PGD^{20}$) results, all defense methods achieve much better robustness against black-box attacks. Compared with other models, ours is capable of generating stronger adversarial examples. Moreover, our model exhibits high robustness and natural accuracy on both datasets, demonstrating our method's excellent performance. For a fair comparison, we keep the training and evaluation settings the same as in MART \cite{c1} and HAT \cite{c13}, and we ensure that the important training details involved in different methods are consistent with the original paper. Thus, we assert that MMAT is the closest to the optimal trade-off among these classical defense methods.
\paragraph{\bf{Discussion of MMAT's effectiveness}}In Fig. \ref{fig7}, we collect the logit features of natural examples and find that our learned features have a larger distance between classes while being more clustered within the same class. The more distinguishable feature embedding justifies our improvement of both RA and NA. In particular, the distribution of MMAT is similar to that of SAT and MART, whose robustness is great, and MMAT can well distinguish ten classes' features of the dataset, like LBGAT, which has high natural accuracy. Then we study the distributions of examples' margins obtained by multiple models. Fig. \ref{fig8} exhibits the distributions. We observe that the margins obtained by our models are consistently in the medium range of all models at different margin levels, in line with our quest for moderate margins. In addition, the example size with small margins of SAT significantly exceed those of MMAT, while the example size with larger margins gradually less than us, which explains why our method can improve both robust accuracy and natural accuracy over SAT. Another interesting finding is that TRADES consistently has the lowest example size when the margin is small, and the highest example size as the margin gradually increases, which makes us understand better why it can stand out under the strong attacks of $\rm CW_\infty$ and AA.
\section{Conclusion}
Based on previous work and our experimental findings, we discover that adversarially trained models cause an excessive increase in the margin along adversarial directions. This partly contributes to the much-debated RA-NA trade-off. The commonly used uniform perturbation budget, $\epsilon$, which ignores the characteristics of examples and leads to the severe issue of cross-over mixture, is a major contributor. To address the issue, our proposed MMAT considers the multiple characteristics of training examples and is fine-tuned by a teacher model to obtain a moderately inclusive decision boundary. Experiments show MMAT achieves a superior RA-NA trade-off compared to existing defenses.
\begin{figure}[t]
\setlength{\abovecaptionskip}{-0.3cm}
\setlength{\belowcaptionskip}{-0.6cm}
\centering
\subfigure[]{
\begin{minipage}[t]{0.4\linewidth}
\centering
\includegraphics[width=1.5in,trim=0 0 0 0]{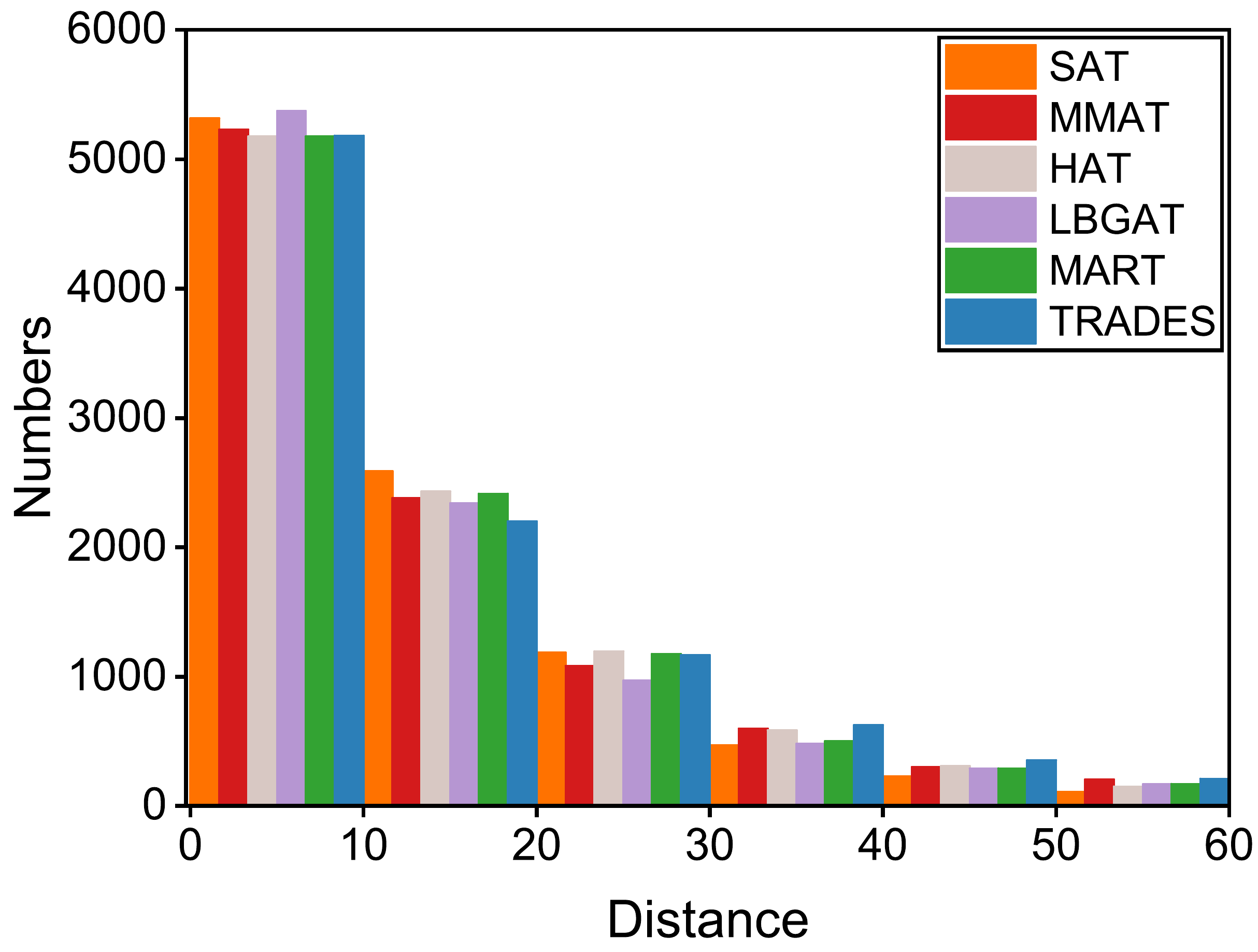}
\end{minipage}
}
\subfigure[]{
\begin{minipage}[t]{0.4\linewidth}
\centering
\includegraphics[width=1.5in,trim=0 0 0 0]{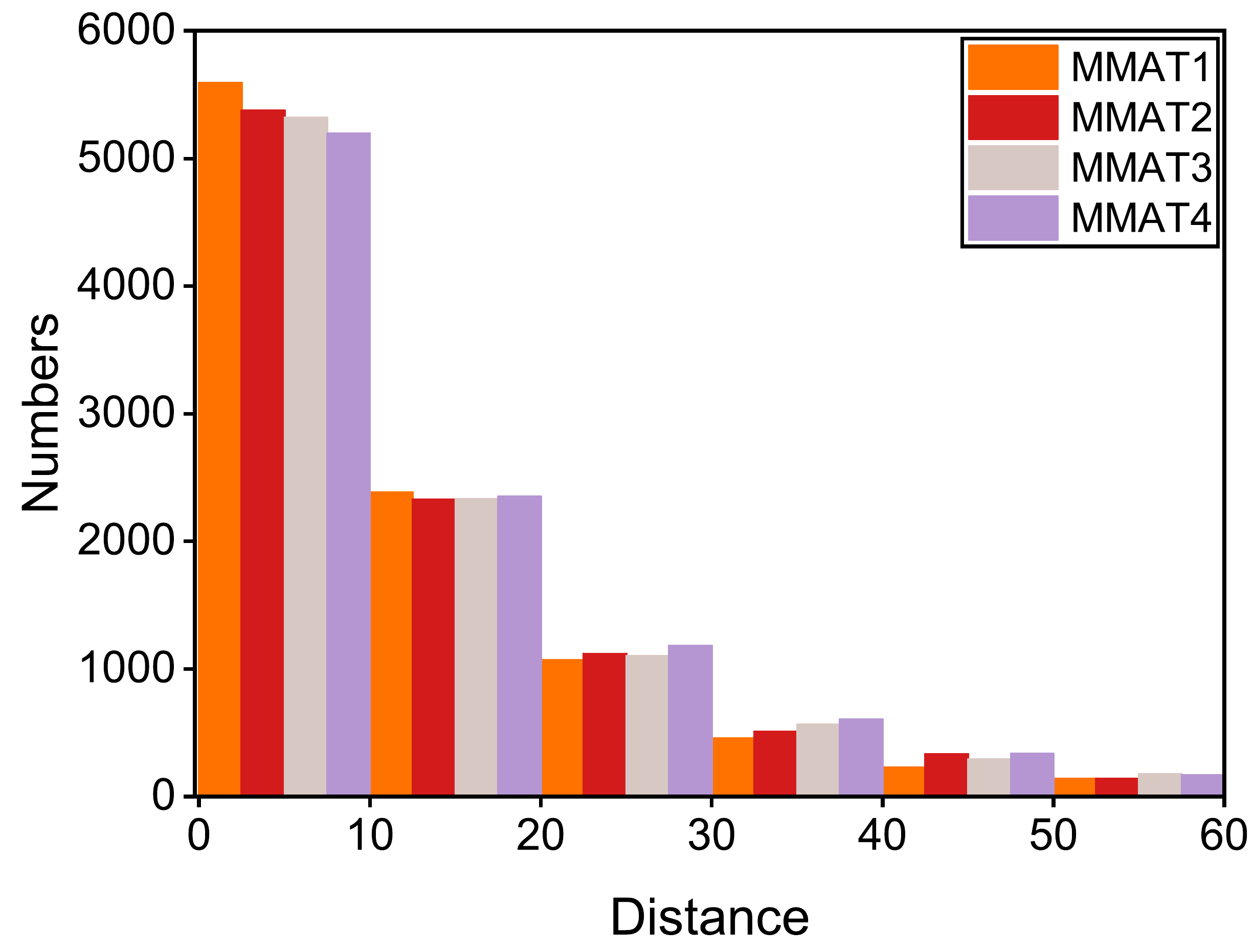}
\end{minipage}
}
\caption{The distribution of examples' margins (the values scaled to 255 times) to the decision boundaries of multiple models. We still use the models trained in Table \ref{tbl2}, and calculate the margin of each model separately. Here we only randomly select 10,000 training examples.}
\label{fig8}
\end{figure}
\section{Acknowledgments}
This work was supported by the Key Program of Zhejiang Provincial Natural Science Foundation of China (LZ22F020007), Major Research Plan of the National Natural Science Foundation of China (92167203), National Key R$\&$D Program of China (2018YFB2100400).

\begin{spacing}{1}
   \bibliographystyle{elsarticle-num}
    \scriptsize
   \bibliography{mybibfile}
\end{spacing}

\end{document}